\documentclass[10pt]{article}

\usepackage{PRIMEarxiv}

\usepackage[utf8]{inputenc} 
\usepackage[T1]{fontenc}    
\usepackage{hyperref}       
\usepackage{url}            
\usepackage{booktabs}       
\usepackage{amsfonts}       
\usepackage{nicefrac}       
\usepackage{microtype}      
\usepackage{lipsum}
\usepackage{fancyhdr}       
\usepackage{graphicx}       
\graphicspath{{media/}}     

\usepackage{lineno,hyperref}
\hypersetup{colorlinks, citecolor=blue, linkcolor=red, urlcolor=Blue}
\usepackage{amsmath}
\usepackage{amsfonts}
\usepackage{amssymb}
\usepackage{geometry}
\usepackage{graphics}
\usepackage{bm}
\usepackage{booktabs}
\usepackage{multirow}
\usepackage{tabularx}
\newcolumntype{Y}{>{\centering\arraybackslash}X}
\usepackage{arydshln}
\usepackage{caption}
\usepackage{subcaption}
\usepackage{threeparttable}
\usepackage{algorithm}
\usepackage[skip=1pt]{caption}
\usepackage{algpseudocode}

\allowdisplaybreaks

\makeatletter
\newcommand{\Note}{\@ifstar{\item[\textbf{Note:}]}{\item[\textbf{Note:}]}}
\makeatother
\usepackage{hyperref}

\newcommand{\Eqref}[1]{\hyperref[#1]{Eq.~(\ref*{#1})}}

\title{A Bayesian Approach for Discovering Time- Delayed Differential Equation from Data
}

\author{
    Debangshu Chowdhury\\
  Department of Applied Mechanics, IIT Delhi \\
  Department of Mechanical Engineering, BITS Pilani, Goa \\
  Goa, India \\
  \texttt{f20210264@goa.bits-pilani.ac.in} \\
   \And
  Souvik Chakraborty \\
  Department of Applied Mechanics \\
  Yardi School of Artificial Intelligence (ScAI) \\
  Indian Institute of Technology Delhi\\
  Hauz Khas, 110016, India\\
  \texttt{souvik@am.iitd.ac.in} \\
}

\begin{document}
\maketitle

\begin{abstract}
Time-delayed differential equations (TDDEs) are widely used to model complex dynamic systems where future states depend on past states with a delay. However, inferring the underlying TDDEs from observed data remains a challenging problem due to the inherent nonlinearity, uncertainty, and noise in real-world systems.
Conventional equation discovery methods often exhibit limitations when dealing with large time delays, relying on deterministic techniques or optimization-based approaches that may struggle with scalability and robustness.
In this paper, we present BayTiDe - \textbf{Bay}esian Approach for Discovering \textbf{Ti}me- \textbf{De}layed Differential Equations from Data, that is capable of identifying arbitrarily large values of time delay to an accuracy that is directly proportional to the resolution of the data input to it.
BayTiDe leverages Bayesian inference combined with a sparsity-promoting discontinuous spike-and-slab prior to accurately identify time-delayed differential equations. The approach accommodates arbitrarily large time delays with accuracy proportional to the input data resolution, while efficiently narrowing the search space to achieve significant computational savings.
We demonstrate the efficiency and robustness of BayTiDe through a range of numerical examples, validating its ability to recover delayed differential equations from noisy data.
\end{abstract}


\section{Introduction}\label{sec:introduction}
Time-delayed differential equations (TDDEs) are crucial for modeling a wide variety of real-world phenomena where the future state of a system depends not only on its present state but also on past states with a time lag. These systems are prevalent in fields such as neuroscience \cite{neurosciencetimedelay}, epidemiology \cite{timedelayepidemiology}, economics \cite{economicstimedelay}, and engineering \cite{engineeringtimedelay}, where delays in feedback loops or information propagation are integral to the system’s dynamics. However, discovering the structure and parameters of TDDEs directly from observed data poses significant challenges because of the non-linear nature of the equations, the complexity of delay terms, and the presence of noise and uncertainty in real-world measurements. As a result, there is a clear need for advanced algorithms capable of discovering TDDEs from data. Moreover, it is important that these algorithms also quantify epistemic uncertainty, accounting for the limited and noisy nature of the measurements used in the equation discovery process.

The process of constructing a mathematical model for an observed system with unknown governing equations generally begins by assuming a functional form for the system's equations and estimating the corresponding parameters using observed data \cite{schmidt2009distilling}. This traditional approach often relies on a priori knowledge of the system’s structure, which can limit its applicability when the system’s behavior is not fully understood. In contrast, an alternative approach involves using data-driven techniques, such as machine learning, to learn the input-output relationship directly from the data \cite{Karniadakis2021}. While this method offers flexibility by avoiding assumptions about the system’s functional form, it can result in models that act as ``black boxes'', offering little to no interpretability or insight into the underlying physics or mechanisms of the system. Furthermore, purely data-driven models often fail to generalize beyond the training dataset, making them unreliable for predicting new, unseen scenarios.
To address these limitations, recent advancements have focused on data-driven equation discovery methods, which aim to derive both the structure and parameters of governing equations directly from data while maintaining interpretability. This task closely resembles model selection problems, where the goal is to select the best model from a set of candidates, balancing the model’s complexity against its ability to explain the data. Traditional model selection approaches, such as those based on the Akaike Information Criterion (AIC) and Bayesian Information Criterion (BIC), rely on navigating the bias-variance trade-off to choose an optimal model \cite{1100705, schwarz1978estimating}. These methods, however, often require assumptions about the model form or may not be applicable to systems with complex, unknown dynamics.
The seminal work in data-driven equation discovery was introduced by symbolic regression and genetic programming techniques \cite{symbolic_regression, genetic-programming}. These methods aim to automatically generate the structure of governing equations—such as ordinary differential equations—by exploring a vast space of possible models. Symbolic regression not only identifies the most appropriate functional form of the system's governing equations but also estimates the associated parameters from the data. This approach has shown significant promise in discovering mathematical models for a wide range of systems, making it a powerful tool for equation discovery, especially when prior knowledge about the system's form is limited or unavailable. However, this algorithm is computationally expensive and does not scale to high-dimensional systems.

To improve the computational efficiency and scalability with respect to data, Sparse Identification of Nonlinear Dynamics was proposed for the discovery of the governing physics of nonlinear dynamical systems \cite{SINDy}. Computational efficiency was improved by limiting the model selection search space to a library of candidate functions and by exploiting the fact that governing equations tend to have only a few candidate functions, i.e, they are sparse. The algorithm enforced sparsity by using sequentially thresholded least squares to select the candidate functions that best fit the data. The function library allows for the implementation of some pre-existing knowledge about the system to be identified. It has since been extended to a variety of applications like the sparse identification of stochastic dynamic equations \cite{boninsegna2018sparse}, identification of biological networks \cite{Mangan201652}, identification of time-varying aerodynamics of a prototype bridge \cite{bridge_sindy}, identification of partial differential equations using a weak form formulation \cite{weak-sindy}, sparse identification of nonlinear vector-valued Ansatz functions \cite{reactive-sindy},  and Data-driven Discovery of Delay Differential Equations with Discrete Delays \cite{time_delay_discrete_discovery}, among others. Independently, an equation discovery framework based on moving horizon optimization was developed that showed remarkable robustness to noise by evaluating sequential subsets of data \cite{mho-discovery}.

The frameworks discussed thus far are deterministic in nature, and were not able to quantify the uncertainty associated with discovering the equations from the library of candidate functions. As such, Bayesian frameworks for the discovery of governing equations from data was proposed \cite{FUENTES2021107528, nayek2021spike, gupta2021bayesian, gibbs-stochastic, nayek2022equation}. The least-squares based linear regression model was replaced by a Bayesian linear regression technique. It functions similarly to Bayesian variable selection \cite{bayesian_variable_selection}, and utilizes Bayesian Inference to probabilistically set the correct model and estimate the parameters, thus eliminating the possibility of overfitting to the input data and exhibiting robustness to noise \cite{gibbs-variable-selection}. Sparsity was induced in the solution by choosing appropriate sparsity promoting priors over the weight vector \cite{ss-prior-frequentist, spike-slab-variable-selection, nayek2021spike, hirsh2021sparsifying}. The output of this approach consisted of the posterior distributions of all associated random variables, allowing one to identify the uncertainty in the variables. Sparse regression was later extended to deal with a combinatorially large number of candidate functions using information criteria in \cite{mangan2017model}, thus improving the accuracy and robustness \cite{ROSAFALCO}.

While significant progress has been made in identifying systems governed by ordinary and partial differential equations, much less attention has been given to functional equations such as Delay Differential Equations (DDEs). Early work on parameter estimation for time-delay systems was conducted in \cite{accel_gaussian_process}, where Gaussian Processes were employed to regress over time series data, allowing samples to be drawn without explicitly solving the dynamical system. Subsequently, \cite{bio-distributed-delays} utilized Bayesian inference to estimate the appropriate time delay for a given model. Extensions of SINDy (Sparse Identification of Nonlinear Dynamics) have also been explored for DDE identification. In \cite{WU2023133647}, the function library was parameterized to include undetermined variables, formulating the problem as a mixed-integer nonlinear programming challenge. Additional SINDy-based approaches have been applied to delay systems in contexts such as directional drilling in oilfield operations \cite{time_delay_gap_effects_sindy} and modeling bacterial zinc response \cite{Sandoz_2023}, using a greedy search over a predefined set of possible delay values. Another line of work used Taylor expansions to parameterize the time delay from the transfer function of second-order nonlinear DDEs \cite{Leylaz_2021}. This approach assumes small time delays and linear delay terms while focusing on identifying the nonlinear terms present in the equations. However, these assumptions significantly limit the applicability of the method to generalized time-delayed systems.
More recent advancements in SINDy involve incorporating Bayesian optimization to iteratively determine the time delay while evaluating errors at each step \cite{time_delay_discrete_discovery}. Similarly, \cite{data-driven-augment} identified time delays by iterating through a candidate set of delay values. Despite these developments, most existing methods perform well only for small time delays and are sensitive to measurement noise. Their effectiveness further diminishes as the nonlinearities of the input system increase. Additionally, these frameworks often require manual tuning of hyperparameters, limiting their robustness and applicability to diverse and complex systems.

In response to the challenges outlined above, we propose \textbf{BayTiDe}, a \textbf{Bay}esian Approach for Discovering \textbf{Ti}me-\textbf{De}layed Differential Equations from Data. BayTiDe leverages sparsity-promoting priors in conjunction with Bayesian inference to systematically identify delay-differential equations from observed data. This approach introduces several distinctive features that address existing limitations in the field:  

\begin{itemize}
    \item \textbf{Uncertainty Quantification:}  
    A fundamental advantage of BayTiDe is its Bayesian formulation, which provides not just point estimates of the governing equations but also the posterior distribution of the weight vector. This allows for the computation of confidence intervals, quantifying the uncertainty in the identified equations. Such an approach is critical for real-world applications where data is noisy or incomplete, as it offers insights into the reliability of the discovered model.  

    \item \textbf{Robustness:}  
    BayTiDe exhibits strong robustness to noise and can handle highly corrupted data without compromising its performance. Moreover, it demonstrates excellent generalization capabilities, enabling accurate predictions even for test data with initial conditions that deviate significantly from those in the training set. Additionally, the framework can generalize to dynamic poles or system behaviors not explicitly captured in the training data, making it suitable for a broad range of applications.  

    \item \textbf{Adaptability to Large Time Delays:}  
    Unlike many existing methods that are limited to identifying small time delays, BayTiDe is capable of handling arbitrarily large delays. Its adaptive mechanism ensures that the time delay is accurately identified irrespective of its magnitude, thus broadening its applicability to systems with extensive lag periods.  

    \item \textbf{Insensitivity to Stability:}  
    BayTiDe is equally effective in identifying systems with varying stability characteristics. It can successfully discover equations governing both periodic systems, which exhibit stable oscillatory behavior, and chaotic systems, which are highly sensitive to initial conditions and exhibit irregular, aperiodic dynamics. This insensitivity to stability makes BayTiDe a versatile tool for analyzing a wide range of dynamic systems.  
\end{itemize}
Overall, BayTiDe provides a robust, reliable, and adaptive framework for the discovery of time-delayed differential equations.

The remainder of the paper is organized as follows: Section \ref{sec:problem formulation} formulates the problem statement of the presented work. Section \ref{sec:methodology} discusses the entire proposed data-driven Bayesian framework and presents the algorithm for the same. Section \ref{sec:Numerical Studies} discusses the numerical studies conducted to demonstrate the capabilities of the developed framework. Finally, in Section \ref{sec:conslusion} the novel contributions of BayTiDe are summarized and the paper is concluded.

\section{Problem Formulation}\label{sec:problem formulation}
In this section we formally define the problem statement by considering a nonlinear delay system. We restrict ourselves to first order systems that have a single, constant time delay that is unknown. Such a system can be represented in general by the Delay Differential Equation (DDE),
\begin{equation}
    {\bm{\dot{X}}} = {F(\\ \bm{X}(t), \bm{X_{\tau}}(t))}, \label{eq:problem formulation}
\end{equation}
where the state vector ${\bm{X}(t) \in \mathbb{R}^m}$ represents the state variables of the system at a time ${t}$, the delayed state vector ${\bm{X_{\tau}}(t) \in \mathbb{R}^m}$ represents the value of the state variables at a time ${(t-\tau)}$, and ${F}$ represents the function(s) of $\bm X$ and $\bm X_\tau$ that determine the system dynamics. The measured data is assumed to be noisy. With this setup, the objective of this paper is to develop a framework that can determine the governing DDE of the system from noisy measurements of the state variables across $n$ time steps. Let this state data be represented by $\mathbf{X}\in \mathbb{R}^{n\times m}$.

Let \( N \) denote the total number of measurements. We assume that the target function \( F \) can be expressed as a linear combination of unknown candidate functions drawn from the set \( \{f_k(\bm{X}, \bm{X_\tau}) : k = 1, \dots, K\} \). For simplicity, we assume that \( f_k(\bm{X}, \bm{X_\tau}) \) represents functions that may depend on a single state vector \( \bm{X} \), the delayed state vector \( \bm{X_\tau} \), or both. 
Let \( \mathbf{L} \in \mathbb{R}^{N \times K} \) denote the library matrix of candidate functions constructed from the set \( \{f_k\} \). The candidate functions may consist of various functional forms that are not necessarily orthogonal to one another. Under this formulation, \( F \) can be represented as a linear combination of these functions:
\begin{equation}\label{eq:basis LC}
    F = \theta_1 f_1 + \theta_2 f_2 + \dots + \theta_k f_k + \dots + \theta_K f_K,
\end{equation}
where \( \theta_k \) are the weights associated with each candidate function \( f_k \). 
To ensure this relationship holds across all \( N \) time steps, the task reduces to solving a regression problem of the form:
\begin{equation}\label{eq:regression problem}
    \bm{Y} = \mathbf{L} \bm{\theta} + \epsilon,
\end{equation}
where \( \bm{Y} \in \mathbb{R}^N \) is the target vector representing the observed measurements, \( \mathbf{L} \in \mathbb{R}^{N \times K} \) is the library matrix of candidate functions, \( \bm{\theta} \in \mathbb{R}^K \) is the weight vector, and \( \epsilon \) accounts for observational noise. 
Therefore, the objective reduces to identifying the relevant candidate functions (model selection) and the associated parameters (parameter estimation). As an added layer of complexity with DDE, the
delay term is also unknown and needs to be estimated.


\section{Proposed approach}\label{sec:methodology}
The regression problem underlying the discovery of time-delayed differential equations involves two key tasks: (1) \textbf{Model Selection} and (2) \textbf{Parameter Estimation}. To address these tasks simultaneously, we employ Sparse Bayesian Linear Regression, leveraging the inherent sparsity of dynamical systems. Dynamical systems are typically governed by a small subset of functions, making them sparse within the high-dimensional function space. Sparsity is induced by assigning specific priors to the weight vector \( \bm{\theta} \), such as the Spike-and-Slab (SS) prior \cite{spike-slab-variable-selection}, which promotes sparsity by shrinking most weights to values close to zero. 
The SS prior consists of two components: a ``spike,'' which concentrates most of the probability density around zero, and a ``slab,'' which diffuses the remaining density over a broader range. In this work, we adopt a variant known as the Discontinuous Spike-and-Slab (DSS) prior, which models the spike as a Dirac delta function \cite{nayek2021spike}. Unlike the traditional SS prior, the DSS prior explicitly sets weights assigned to the spike exactly to zero, rather than to small values near zero. This property accelerates convergence by allowing these weights to be excluded from the sampling process, as they no longer contribute to model selection. 

\begin{figure}[ht]
    \centering
    \includegraphics[width=\linewidth]{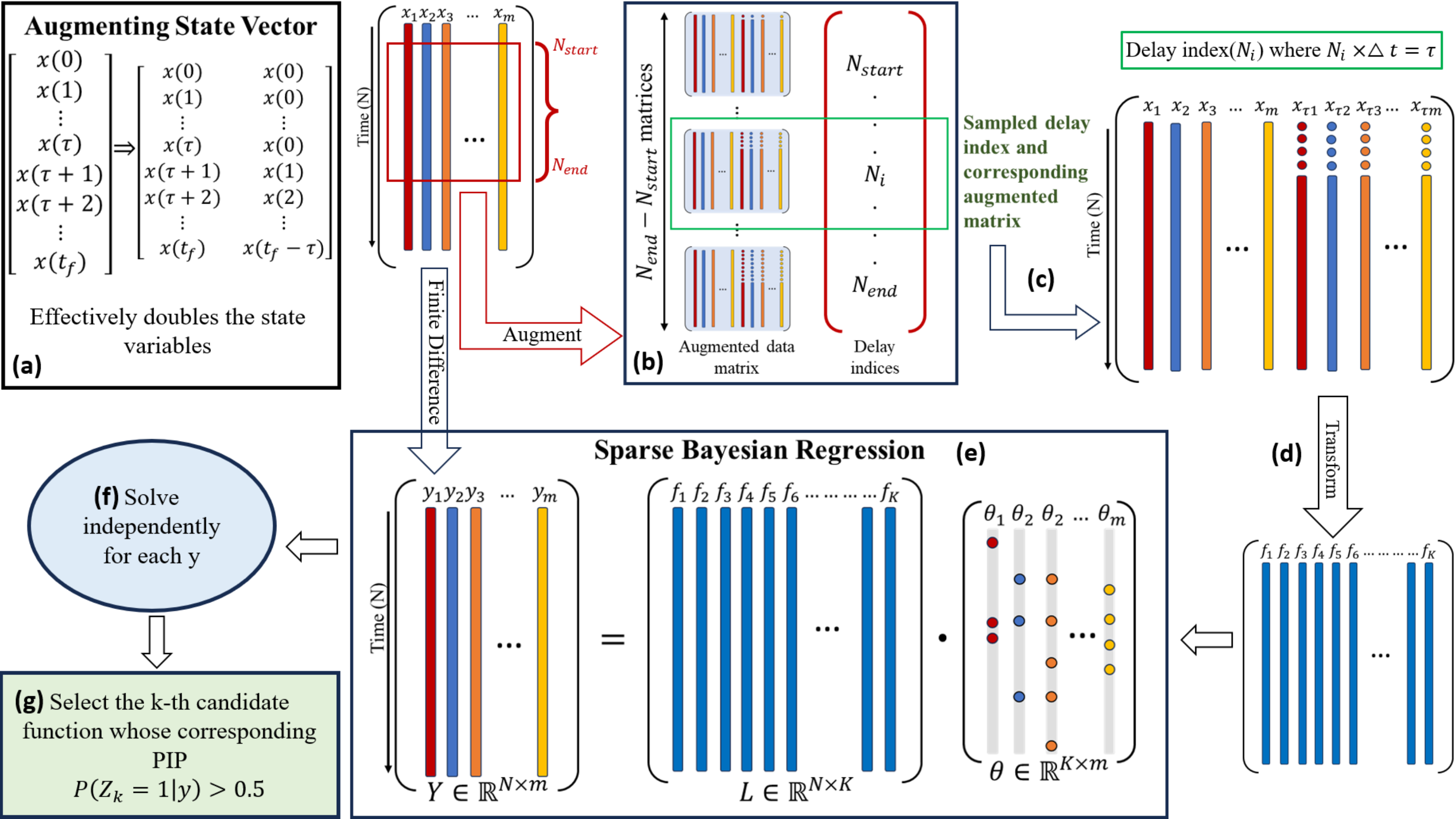}
    \caption{\textbf{Schematic illustration:} BayTiDe is the combination of two key ideas and sparse Bayesian regression. The first idea is the augmenting of the state data with artificial state variables that correspond to the delay terms within the governing equations \textbf{(a)}. This effectively doubles the number of variables to be identifies but allows for the inclusion of functions of two or more delay terms. The delay index has to be sampled from an arbitrarily large search space of indices ($N_{start},N_{end}$), each with a corresponding unique augmented data matrix \textbf{(b)}. The second key idea is to model this sampling using a Multinomial distribution, assigning each index with a finite probability of being chosen. The framework samples the delay index and the corresponding augmented data matrix \textbf{(c)}. The matrix is then transformed into the library $\mathbf{L}$ using the candidate functions $\{f_1, f_2, ..., f_K\}$ \textbf{(d)}. This library is utilized for performing sparse Bayesian linear regression. The candidate functions in the library are parameterized by a weight vector $\bm \theta$ whose sparsity (shown by the dots in the figure) is promoted through the use of sparsity promoting priors \textbf{(e)}. Each element of the weight vector is assigned a latent variable $Z_k: k=1,2,...K$ to classify the weight as a spike or a slab. The framework is run independently for each state derivative $y_i$ to find the entire system of DDEs \textbf{(f)}. After the Bayesian regression is completed, only the functions whose corresponding PIP ($P(Z_K=1|Y)$ is greater than 0.5 are considered in the final predicted model \textbf{(g)}.}
    \label{fig:schematic}
\end{figure}

To construct the regression library \( \mathbf{L} \), the time delay must first be determined. In our framework, both the time delay and the library are treated as random variables to be inferred from the data. The library is parameterized alongside the time delay, but this parameterization lacks an explicit functional form, rendering the joint posterior distribution intractable. To address this challenge, we employ Gibbs sampling, iteratively drawing samples from the conditional distributions of the model parameters. A schematic representation of the overall BayTiDe framework is shown in \autoref{fig:schematic}. 
The following section formally introduces the BayTiDe methodology, detailing the parameterization process and deriving the conditional sampling distributions for all random variables involved. This approach ensures a principled inference framework for simultaneous model selection and parameter estimation in time-delayed dynamical systems. 

\subsection{Prior Distributions}\label{subsec:augment}
To parameterize the library and the time delay, we introduce the concept of augmenting the data matrix. For each state variable in the system to be identified, we generate auxiliary variables that represent the delayed states, denoted by \( \mathbf{X_\tau} \) in \Eqref{eq:problem formulation}. These auxiliary variables are then concatenated with the original state variables to construct an augmented data matrix (refer to \autoref{fig:schematic}(a) \cite{data-driven-augment}). This augmentation enables the incorporation of delay terms into the regression framework. 
Since the augmentation process discretizes the time delay into multiples of \( \Delta t \), the resolution of the observed data, we reformulate the problem by introducing the concept of a \textbf{time delay index} \( \tau \). Throughout this section, \( \tau \) refers to the delay index and not the actual time delay. The relationship between the two is given by  
\(
\tau \times \Delta t = \text{time delay}.
\)  
This reformulation simplifies the problem to estimating \( \tau \), which can take integer values in the range \( [0, N-1] \), where \( N \) is the total number of time steps.  
We treat \( \tau \) as the outcome of a multinomial trial, where each integer in the range \( [0, N-1] \) represents a possible outcome. This approach is analogous to rolling an \( N \)-sided die to determine \( \tau \). The probabilities associated with each outcome, denoted as \( g_i \) for \( i = 0, 1, \ldots, N-1 \), are also treated as random variables. To enable probabilistic inference, we assign a Dirichlet prior to these probabilities, leveraging its property as a conjugate prior for the multinomial distribution. Once a value of \( \tau \) is sampled, the corresponding library \( \mathbf{L_\tau} \) is constructed, allowing for the discovery of the governing equations.
To summarize, the process is described by the following:
\begin{subequations}
    \begin{align}
        \tau &\sim \text{Multi}(g_0, g_1, \ldots, g_{N-1}), \\
        g_0, g_1, \ldots, g_{N-1} &\sim \text{Dirichlet}(\alpha_0, \alpha_1, \ldots, \alpha_{N-1}),
    \end{align}
\end{subequations}
where \( \alpha_i \) for \( i = 0, 1, \ldots, N-1 \) are the hyperparameters of the Dirichlet prior. Let \( G \) and \( \alpha_G \) represent the vector of parameters and hyperparameters, respectively. With \( \tau \) and \( \mathbf{L} \) now established, we proceed to define the remaining components of the proposed BayTiDe framework.

In \Eqref{eq:regression problem}, the noise vector \( \epsilon \) is assumed to follow a zero-mean multivariate Gaussian distribution with variance \( \sigma^2 \), i.e., \( \epsilon \sim \mathcal{N}(0, \sigma^2 I_N) \), where \( I_N \) denotes the \( N \times N \) identity matrix. Consequently, the likelihood function is given by:
\begin{equation}\label{eq:y|theta} 
    \bm{Y} | \theta, \mathbf{L_\tau}, \sigma^2 \sim \mathcal{N}(\bm{L_\tau} \theta, \sigma^2 I_N).
\end{equation}
In equation discovery, it is assumed that only a few weights in the weight vector \(\bm \theta \) are non-zero. This sparsity is enforced through the use of DSS-priors. To facilitate this, a latent binary vector \( \bm{Z} = [Z_0, Z_1, \ldots, Z_K] \) is introduced, where each element \( Z_k \) takes the value 1 if the corresponding weight \( \theta_k \) follows a slab distribution (i.e., an active function) and 0 if it follows a spike distribution. The spike distribution is represented by a Dirac-Delta function, which ensures that the corresponding weights do not contribute to the equation discovery process. 
Let \( \bm{\theta_r} \) denote the reduced weight vector, which includes only the weights corresponding to latent variables \( Z_k = 1 \) (i.e., the active functions). Hence, the conditional distribution of \( \bm{\theta} \) given the latent vector \( \bm{Z} \) is:
\begin{equation}\label{eq:spike and slab distribution}
    p(\bm{\theta} | \bm{Z}) = p_{\text{slab}}(\bm{\theta_r}) \prod_{k, Z_k = 0} p_{\text{spike}}(\theta_k),
\end{equation}
where the slab distribution is modeled as a multivariate Gaussian with a variance \( \nu \sigma^2 \), i.e., \( p(\bm{\theta_r} | \bm{Z}) = \mathcal{N}(0, \nu \sigma^2 I_r) \).
$\bm{L_r}$ is the reduced library that only contains r columns whose weights are sampled from the slab distribution. 
The noise variance is scaled by the slab variance to adjust the weights according to the probabilistic variation of the target variable. To enhance robustness to noise and facilitate faster convergence, both the noise variance \( \sigma^2 \) and the slab variance \( \nu \) are treated as random variables. They are assigned Inverse-Gamma (IG) priors, as the IG distribution is a conjugate prior for the Gaussian distribution and ensures that only positive values are sampled. The IG priors are characterized by the hyperparameters \( \alpha_\nu, \beta_\nu \) for \( \nu \) and \( \alpha_\sigma, \beta_\sigma \) for \( \sigma^2 \), respectively.
Each latent variable \( Z_k \), which determines the classification of the corresponding weight, is also treated as a random variable. The classification of the weights can be viewed as a Bernoulli trial with a success probability \( p_0 \). Accordingly, \( p_0 \) is assigned a conjugate Beta prior, with hyperparameters \( \alpha_p \) and \( \beta_p \). The priors for these variables are summarized as follows:
\begin{subequations}\label{eq:priors}
    \begin{align}
        p(\nu) &\sim IG(\alpha_\nu, \beta_\nu), \\
        p(\sigma^2) &\sim IG(\alpha_\sigma, \beta_\sigma), \\
        p(Z_k|p_0) &\sim \text{Bern}(p_0), \\
        p(p_0) &\sim \text{Beta}(\alpha_p, \beta_p).
    \end{align}
\end{subequations}
With these priors defined, a Hierarchical Bayesian model is constructed that enforces conditional independence between certain random variables. Next, details on the Hierarchical Bayesian model are provided.

\subsection{Hierarchical Bayesian Model}\label{subsec:bayesian network}
\begin{figure}
    \centering
    \includegraphics[width=0.5\linewidth]{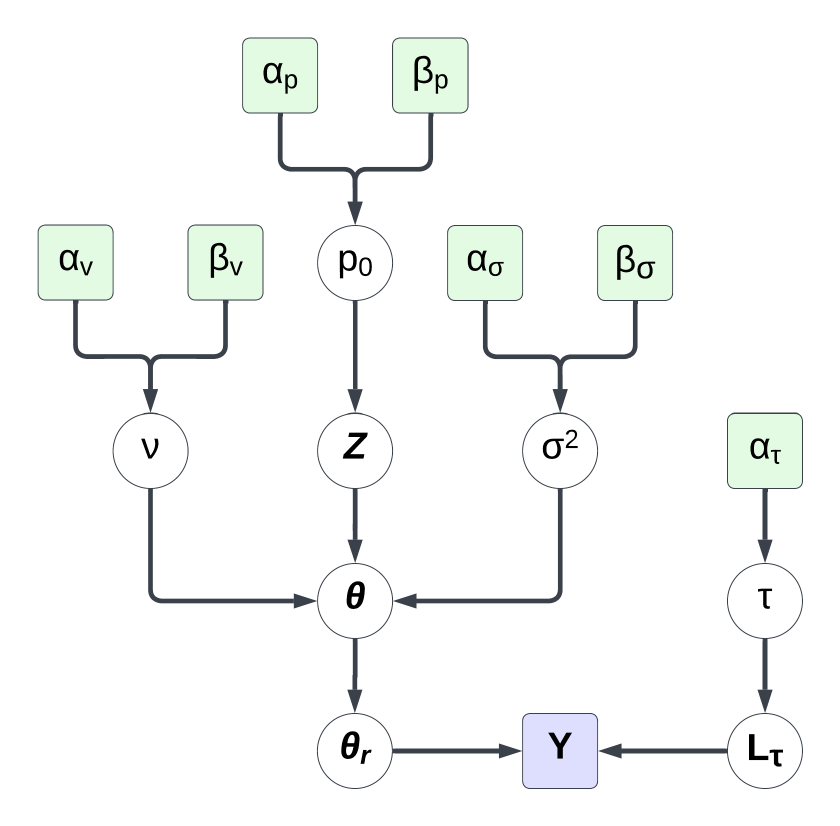}
    \caption{\textbf{Hierarchical Bayesian Network:} Green boxes represent constant hyper-parameters input to the system. White boxes are the random variables that are sampled in each iteration of sampling. $\mathbf{Y}$ represents the first derivative of the state variable and $\mathbf{L_\tau}$ represents a library of candidate functions constructed using the measured data. }
    \label{fig:hierarchical bayesian system}
\end{figure}

Given the prior distributions and setup outlined earlier, a hierarchical Bayesian model is formulated, as illustrated in \autoref{fig:hierarchical bayesian system}. By applying Bayes' Theorem, the posterior probability to be determined is expressed as:
\begin{equation}\label{eq:total posterior}
    p(p_0, \nu, \theta, Z, \tau, \mathbf{L_\tau}, \sigma^2|\textbf{Y}) = \frac{p(\bm{Y}|\theta,\mathbf{L_\tau}, \sigma^2)p(\theta|Z,\nu,\sigma^2)p(Z|p_0)p(p_0)p(\nu)p(\sigma^2)p(\mathbf{L_\tau}|\tau)p(\tau)}{p(\textbf{Y})}.
\end{equation}
Direct sampling from this distribution is not feasible because the marginal likelihood \( p(\mathbf{Y}) \) is intractable due to the DSS priors and the unknown nature of the library. To address this, Gibbs sampling is employed to sequentially draw samples from the conditional distributions of the random variables, conditioned on the others. The directed acyclic graph (DAG) depicted in the figure simplifies some of these dependencies by imposing conditional independence between selected variables. The conditional independence assumptions are as follows:
\begin{subequations}
    \begin{align}
        p(p_0|\mathbf{Y}, \bm \theta, \mathbf{L_\tau}, \sigma^2, Z, \nu, \tau, G) &= p(p_0|Z) \\
        p(\nu|\mathbf{Y}, \bm \theta, \mathbf{L_\tau}, \sigma^2, Z, p_0, \tau, G) &= p(\nu|\bm \theta, \sigma^2, Z) \\
        p(\sigma^2|\mathbf{Y}, \bm \theta, \mathbf{L_\tau}, p_0, Z, \nu, \tau, G) &= p(\sigma^2|\mathbf{Y}, \bm \theta, \mathbf{L_\tau}, Z, \nu) \\
        p(Z|\mathbf{Y}, \bm \theta, \mathbf{L_\tau}, \sigma^2, p_0, \nu, \tau, G) &= p(Z|\mathbf{Y}, \bm \theta, \mathbf{L_\tau}, \sigma^2, p_0) \\
        p(\bm \theta|\mathbf{Y}, p_0, \mathbf{L_\tau}, \sigma^2, Z, \nu, \tau, G) &= p(\bm \theta|\mathbf{Y}, \mathbf{L_\tau}, \sigma^2, Z, \nu) \\
        p(\tau|\mathbf{Y}, \bm \theta, \mathbf{L_\tau}, \sigma^2, Z, \nu, p_0, G) &= p(\tau|\mathbf{L_\tau}, \mathbf{Y}, \bm \theta, \sigma^2, Z, \nu) \\
        p(\mathbf{L_\tau}|\mathbf{Y}, \bm \theta, p_0, \sigma^2, Z, \nu, \tau, G) &= p(\mathbf{L_\tau}|\tau) \\
        p(G|\mathbf{L_\tau}, \mathbf{Y}, \bm \theta, p_0, \sigma^2, Z, \nu, \tau)  &= p(G|\tau)
    \end{align}
\end{subequations}
It is important to note that certain variables are conditioned on \( \mathbf{L_\tau} \), even though the DAG suggests conditional independence in these cases. This approximation is necessary because \( \mathbf{L_\tau} \) lacks an explicit analytical form, which makes the conditional distributions intractable.
With these assumptions, we proceed in the following sequential manner,
\begin{itemize}
\item \textbf{Sampling $p_0$:} The probability of each candidate function being assigned a slab distribution is derived as:  
\begin{subequations}
\begin{align}
    p(p_0|Z) &\propto p(Z|p_0) p(p_0) \\
    &\propto \prod_{k=1}^{K} p_0^{Z_k}(1-p_0)^{1-Z_k} \cdot p_0^{\alpha_p-1}(1-p_0)^{\beta_p-1} \\
    &\propto p_0^{\alpha_p-1 + h_Z}(1-p_0)^{\beta_p-1 + K - h_Z},
\end{align}
\label{eq:p derivation}
\end{subequations}
where $h_Z = \sum_{k=1}^K Z_k$. The right-hand side (RHS) has the form of a Beta distribution. Thus, $p_0$ can be sampled as:  
\begin{equation}
p_0|Z \sim \text{Beta}(\alpha_p + h_Z, \beta_p + K - h_Z).
\end{equation}

\item \textbf{Sampling $\nu$:} The variance of the slab is derived as follows:  
\begin{subequations}\label{eq:nu derivation}
\begin{align}
    p(\nu|\bm{\theta}, \sigma^2, Z) &\propto p(\bm{\theta}|\nu, \sigma^2, Z)p(\nu) \\
    &\propto p(\bm{\theta}_r|0, \nu\sigma^2I_r) \cdot IG(\nu|\alpha_\nu, \beta_\nu) \\
    &\propto \nu^{-r/2} \exp\left(-\frac{\bm{\theta}_r^T\bm{\theta}_r}{2\nu\sigma^2}\right) \cdot \nu^{-\alpha_\nu-1} \exp\left(-\frac{\beta_\nu}{\nu}\right) \\
    &\propto \nu^{-(\alpha_\nu + r/2)-1} \exp\left(-\frac{\beta_\nu + \frac{\bm{\theta}_r^T\bm{\theta}_r}{2\sigma^2}}{\nu}\right).
\end{align}
\label{eq:nu_derivation}
\end{subequations}  
The final form matches an Inverse Gamma distribution. Therefore, $\nu$ can be sampled as:  
\begin{equation}
\nu|Z, \bm{\theta}, \sigma^2 \sim IG\left(\alpha_\nu + \frac{r}{2}, \beta_\nu + \frac{\bm{\theta}_r^T\bm{\theta}_r}{2\sigma^2}\right).
\end{equation}

\item \textbf{Sampling $\bm{\theta}$:} The weight vector is given by:  
\begin{subequations}\label{eq:theta deriv}
\begin{align}
    p(\bm{\theta}|\mathbf{Y}, \mathbf{L}_\tau, \sigma^2, Z, \nu) &\propto p(\mathbf{Y}|\mathbf{L}_{r,\tau}, Z, \sigma^2) p(\bm{\theta}_r|Z, \nu, \sigma^2) \\
    &\propto N(\mathbf{Y}|\mathbf{L}_{r,\tau}\bm{\theta}_r, \sigma^2I_N) \cdot N(\bm{\theta}_r|0, \nu\sigma^2I_r) \\
    &\propto \exp\left(-\frac{(\mathbf{Y} - \mathbf{L}_{r,\tau}\bm{\theta}_r)^T(\mathbf{Y} - \mathbf{L}_{r,\tau}\bm{\theta}_r)}{2\sigma^2}\right) \cdot \exp\left(-\frac{\bm{\theta}_r^T\bm{\theta}_r}{2\nu\sigma^2}\right) \\
    &\propto \exp\left(-\frac{1}{2\sigma^2} \left(\mathbf{Y}^T\mathbf{Y} - 2\mathbf{Y}^T\mathbf{L}_{r,\tau}\bm{\theta}_r + \bm{\theta}_r^T\mathbf{L}_{r,\tau}^T\mathbf{L}_{r,\tau}\bm{\theta}_r + \nu^{-1}\bm{\theta}_r^T\bm{\theta}_r\right)\right) \\
    &\propto \exp\left(-\frac{1}{2\sigma^2} (\bm{\theta}_r - \mu)^T\Sigma^{-1} (\bm{\theta}_r - \mu)\right),
\end{align}
\label{eq:theta_derivation}
\end{subequations}  
where $\Sigma^{-1} = \mathbf{L}_{r,\tau}^T\mathbf{L}_{r,\tau} + \nu^{-1}I_r$ and $\mu = \Sigma \mathbf{L}_{r,\tau}^T \mathbf{Y}$. Comparing to the standard form of a multivariate Gaussian, $\bm{\theta}_r$ can be sampled as:  
\begin{equation}
\bm{\theta}_r|\mathbf{Y}, \mathbf{L}_{r,\tau}, \nu, \sigma^2 \sim N(\mu, \sigma^2\Sigma).
\end{equation}

\item \textbf{Sampling $\sigma^2$:}  
Gibbs sampling necessitates that the Markov chain is irreducible to ensure eventual convergence to the stationary distribution. As $Z_k = 0, \forall k = 1, 2, \ldots, K$ represents an absorbing state, the parameter $\bm{\theta}$ must be marginalized out of the conditional distribution for $\sigma^2$. This is expressed as:  
\begin{subequations}
    \begin{align}
        p(\sigma^2|\mathbf{Y}, \mathbf{L_\tau}, Z, \nu) &= \int p\left(\sigma^2, \bm{\theta}|\mathbf{Y}, \mathbf{L_\tau}, Z, \nu\right) \, d\bm{\theta} \nonumber \\
        &\propto \int p\left(\mathbf{Y}|\sigma^2, \bm{\theta}, \mathbf{L_\tau}, Z, \nu\right)p\left(\bm{\theta}|Z, \nu, \sigma^2\right)p\left(\sigma^2\right) \, d\bm{\theta} \nonumber \\
        &\propto \left(\int N\left(\mathbf{Y}|\mathbf{L_{r,\tau}}\bm{\theta}_r, \sigma^2I_N\right)N\left(\bm{\theta}_r|0, \nu\sigma^2I_r\right) \, d\bm{\theta}_r \right)p(\sigma^2). \nonumber
    \end{align}
\end{subequations}
The integral term can be expressed using the parameters $\mu$ and $\Sigma$ (defined in Equation~\eqref{eq:theta deriv}) as follows:  
\begin{subequations}\label{eq:sigma2 derivation}
\begin{align}
    p(\sigma^2|\mathbf{Y}, \mathbf{L_\tau}, Z, \nu) &\propto \frac{1}{(2\pi\sigma^2)^{N/2}} \cdot \frac{1}{(2\pi\nu\sigma^2)^{r/2}} \cdot \left(\int N(\bm{\theta}_r|\mu, \sigma^2\Sigma) \, d\bm{\theta}_r\right) \cdot \exp\left(-\frac{\mathbf{Y}^T\mathbf{Y} - \mu^T\Sigma^{-1}\mu}{2\sigma^2}\right)p(\sigma^2) \\
    &\propto \frac{1}{(\sigma^2)^{(N+r)/2}} \cdot \exp\left(-\frac{\mathbf{Y}^T\mathbf{Y} - \mu^T\Sigma^{-1}\mu}{2\sigma^2}\right)p(\sigma^2) \\
    &\propto \frac{1}{(\sigma^2)^{(N+r)/2}} \cdot \exp\left(-\frac{\mathbf{Y}^T\mathbf{Y} - \mu^T\Sigma^{-1}\mu}{2\sigma^2}\right) (\sigma^2)^{-\alpha_\sigma - 1} \cdot \exp\left(-\frac{\beta_\sigma}{\sigma^2}\right) \\
    &\propto (\sigma^2)^{-\left(\frac{r+N}{2} + \alpha_\sigma\right) - 1} \cdot \exp\left(-\frac{1}{\sigma^2}\left(\frac{\mathbf{Y}^T\mathbf{Y} - \mu^T\Sigma^{-1}\mu}{2} + \beta_\sigma\right)\right).
\end{align}
\end{subequations}
From the resulting expression, it is evident that $\sigma^2$ follows an Inverse Gamma distribution. Thus, $\sigma^2$ is sampled as:  
\begin{equation}
    \sigma^2|\mathbf{Y}, \mathbf{L_\tau}, Z, \nu \sim IG\left(\alpha_\sigma + \frac{r+N}{2}, \beta_\sigma + \frac{\mathbf{Y}^T\mathbf{Y} - \mu^T\Sigma^{-1}\mu}{2}\right).
\end{equation}

\item \textbf{Sampling $Z_k$:}  
Each latent variable within the latent vector is sampled individually, conditioned on the values of all other latent variables from the previous sample ($Z_{-k}$). The conditional distribution is estimated by computing the probability of $Z_k=1$ while holding $Z_{-k}$ fixed. Denoting this probability by $\mu_k$, it is expressed as:  
\begin{align}
    \mu_k &\;=\; \frac{p\left( Z_k=1| \mathbf{Y}, \mathbf{L_\tau}, Z_{-k}, \nu, p_0 \right)}{p\left( Z_k=1| \mathbf{Y}, \mathbf{L_\tau}, Z_{-k}, \nu, p_0 \right) + p\left( Z_k=0| \mathbf{Y}, \mathbf{L_\tau}, Z_{-k}, \nu, p_0 \right)} \nonumber \\
    &\;=\; \frac{p\left( \mathbf{Y}|Z_k=1, \mathbf{L_\tau}, Z_{-k}, \nu \right)p\left(Z_k=1|p_0\right)}{p\left( \mathbf{Y}|Z_k=1, \mathbf{L_\tau}, Z_{-k}, \nu \right)p\left(Z_k=1|p_0\right) + p\left(\mathbf{Y}|Z_k=0, \mathbf{L_\tau}, Z_{-k}, \nu\right)p\left(Z_k=0|p_0\right)} \nonumber \\
    &\;=\; \frac{p\left( \mathbf{Y}| Z_k=1, \mathbf{L_\tau}, Z_{-k}, \nu \right)p_0}{p\left(\mathbf{Y}|Z_k=1, \mathbf{L_\tau}, Z_{-k}, \nu \right)p_0 + p\left( \mathbf{Y}|Z_k=0, \mathbf{L_\tau}, Z_{-k}, \nu\right)(1-p_0)} \nonumber \\
    &\;=\; \frac{p_0}{p_0 + \lambda_k(1-p_0)}, \quad \text{where}\; \lambda_k = \frac{p\left(\mathbf{Y}|Z_k=0, \mathbf{L_\tau}, Z_{-k}, \nu\right)}{p\left( \mathbf{Y}| Z_k=1, \mathbf{L_\tau}, Z_{-k}, \nu\right)} \label{eq:mu}
\end{align}
The marginal likelihoods comprising the numerator and denominator of $\lambda_k$ are computed by integrating out $\bm \theta$ and $\sigma^2$, while treating $\mathbf{L_\tau}$ as fixed:
\begin{align}
    p\left( \mathbf{Y}|Z, \mathbf{L_\tau}, \nu \right) &\;=\; \int \int p\left( \mathbf{Y}, \sigma^2, \bm \theta | Z, \nu, \mathbf{L_\tau} \right) d\bm \theta d\sigma^2 \nonumber \\
    &\;=\; \int \int p\left( \mathbf{Y}|\bm \theta, \mathbf{L_\tau}, Z, \sigma^2 \right) p\left( \bm \theta|Z, \nu, \sigma^2 \right) p\left( \sigma^2 \right) 
    d\bm \theta d\sigma^2 \nonumber \\
    &\;=\; \int \left( \int N\left(\mathbf{Y}|\mathbf{L_{\tau,r}} \bm \theta_r, \sigma^2 I_N \right) N\left( \bm \theta_r|0, \nu \sigma^2 I_r \right) 
    d\bm \theta \right) p\left(\sigma^2 \right) d\sigma^2 \nonumber \\
    &\;=\; \int \frac{1}{\left(2\pi\sigma^2\right)^{N/2}} \frac{|\Sigma^{-1}|^{1/2}}{(\nu)^{r/2}} \exp\left( \frac{-\left(\mathbf{Y}^T \mathbf{Y} - \mu^T \Sigma^{-1} \mu\right)}{2\sigma^2} \right) p\left(\sigma^2\right) d\sigma^2 \nonumber \\
    &\;=\; \frac{|\Sigma^{-1}|^{1/2}}{(2\pi)^{N/2}} \frac{(\beta_\sigma)^{\alpha_\sigma}}{(\nu)^{r/2}} \frac{\Gamma\left(\alpha_\sigma+\frac{N}{2}\right)}{\Gamma(\alpha_\sigma)} \frac{1}{\left(\beta_\sigma+\frac{1}{2}\left(\mathbf{Y}^T\mathbf{Y}-\mu^T\Sigma^{-1}\mu\right)\right)^{\alpha_\sigma + \frac{N}{2}}} \label{eq:py given znot0}
\end{align}
Here, $\Gamma(\cdot)$ denotes the Gamma function, and $|\cdot|$ represents the determinant. When all $Z_k=0$, the functional form changes to:
\begin{equation}\label{eq:py given z0}
    p\left( \mathbf{Y}|Z=0, \mathbf{L_\tau}, \nu \right) \;=\; \frac{(\beta_\sigma)^{\alpha_\sigma}}{(2\pi)^{N/2}} \frac{\Gamma\left(\alpha_\sigma+\frac{N}{2}\right)}{\Gamma(\alpha_\sigma)} \frac{1}{\left(\beta_\sigma+\frac{1}{2}\mathbf{Y}^T\mathbf{Y}\right)^{\alpha_\sigma + \frac{N}{2}}}
\end{equation}

\item \textbf{Sampling $\tau$:}  
The delay index $\tau$ is sampled similarly to the latent variables, as each possible delay corresponds to an outcome of a single multinomial trial. Let the probability of each outcome be denoted by $\zeta_j$, conditioned on $Z$, $\bm \theta$, $\nu$, and $g_j$ for $j = 0, 1, \dots, N-1$. The probability $\zeta_j$ is given by:  
\begin{equation}
    \zeta_j = \frac{p\left( \tau = j | \mathbf{Y}, \mathbf{L_j}, Z_k, \nu, p_0, G \right)}{\sum\limits_{l=0}^{N-1} p\left( \tau = l | \mathbf{Y}, \mathbf{L_l}, Z_k, \nu, p_0, G \right)},
\end{equation}
where $\mathbf{L_l}$ represents the library augmented at the delay index $l$. Consequently, the library must be re-augmented for each term in the summation within the denominator. Simplifying further, we obtain:
\begin{align}
    \zeta_j &\;=\; \cfrac{p\left( \mathbf{Y} | \tau = j, \mathbf{L_j}, Z_k, \nu, p_0 \right) \; p\left( \tau = j | G \right)}{\sum\limits_{l=0}^{N-1} p\left( \mathbf{Y} | \tau = l, \mathbf{L_l}, Z_k, \nu, p_0 \right) p\left( \tau = l | G \right)} \nonumber \\
    &\;=\; \cfrac{p\left( \mathbf{Y} | \tau = j, \mathbf{L_j}, Z_k, \nu, p_0 \right) \; g_j}{\sum\limits_{l=0}^{N-1} p\left( \mathbf{Y} | \tau = l, \mathbf{L_l}, Z_k, \nu, p_0 \right) g_l} \nonumber \\
    &\;=\; \frac{g_j}{g_j \;+\; \sum\limits_{l=0, l \neq j}^{N-1} \omega_l g_l}, \label{eq:zeta derivation}
\end{align}
where $\omega_l = \frac{p\left( \mathbf{Y} | \tau = l, \mathbf{L_l}, Z_k, \nu, p_0 \right)}{p\left( \mathbf{Y} | \tau = j, \mathbf{L_j}, Z_k, \nu, p_0 \right)}$. The quantities in the numerator and denominator can be computed using \Eqref{eq:py given znot0}. Once $\tau$ is sampled, the corresponding library $\mathbf{L_\tau}$ can be constructed.

\item \textbf{Sampling $G$:}  
Given that $\tau$ is sampled only once, the likelihood $p\left( \tau | G \right)$ reduces to the parameter associated with the sampled delay index. Denoting this parameter by $g_j$, the posterior distribution for $G$ can be derived as follows:  
\begin{subequations}\label{eq:G derivation}
    \begin{align}
        p\left( G | \tau \right) &\propto p\left( \tau | G \right) p\left( G \right) \\
        &\propto g_j \prod\limits_{l=0}^{N-1} g_l^{\alpha_l - 1} \\
        &\propto g_j^{(\alpha_j + 1) - 1} \prod\limits_{l=0, l \neq j}^{N-1} g_l^{\alpha_l - 1}.
    \end{align}
\end{subequations}  
Thus, $G$ can be sampled as  
\[
G | \tau \sim \text{Dirichlet}\left( \alpha_0, \alpha_1, \dots, \alpha_{j-1}, \alpha_j + 1, \alpha_{j+1}, \dots, \alpha_{N-1} \right),
\]  
where the hyperparameter $\alpha_j$ associated with the sampled delay index $j$ is incremented by 1, while all other hyperparameters remain unchanged.

\end{itemize}

To summarize, the random variables $\{\tau^t, \mathbf{L_\tau}^t, Z^t, (\sigma^2)^t, \nu^t, p_0^t, G^t, \bm \theta^t \}$ are sampled sequentially using conditional distributions derived above. The superscript $t$ indicates the values sampled in each iteration of the algorithm. 
\begin{enumerate}
    \item The time delay index is sampled from a Multinomial Distribution whose parameters $\zeta_l,\;l=0,1,...N-1$ are calculated using \Eqref{eq:zeta derivation},
    \begin{equation}\label{eq: sampling tau}
         \tau^{(t+1)}|Z^t, \mathbf{Y}, \nu^t, G^t, \mathbf{L_\tau}^t\; \sim \; Multinomial(\mathbf{\zeta})
    \end{equation}
    \item The library $\mathbf{L_\tau}^{t+1}$ is constructed using $\tau^{t+1}$
    \item The latent variables $Z_k^{(t+1)}$ are sampled from the Bernoulli Distribution,
    \begin{equation}\label{eq:sampling Z}
        Z_k^{(t+1)}|\mathbf{Y}, \mathbf{L_\tau}^{(t+1)}, \nu^{(i)},p0^{(t)} \;\sim\; Bern(\mu_k)
    \end{equation}
    where $\mu_k$ is calculated using \Eqref{eq:mu}.
    \item The noise variance $(\sigma^2)^{t+1}$ is sampled from an Inverse Gamma distribution as shown in \Eqref{eq:sigma2 derivation},
    \begin{equation}
        (\sigma^2)^{t+1}|\mathbf{Y}, \mathbf{L_\tau}^{t+1}, Z^{t+1},\nu^t\;\sim\; IG\left(\alpha_\sigma+\frac{r+N}{2}, \beta_\sigma + \frac{\mathbf{Y}^T\mathbf{Y} - \mu^T\Sigma^{-1}\mu}{2} \right) \label{eq:sampling sigma}
    \end{equation}
    \item The slab variance $\nu^{t+1}$ is sampled from an Inverse Gamma distribution as shown in \Eqref{eq:nu derivation},
    \begin{equation}
        \nu^{t+1}|Z^{t+1},\bm \theta^t,(\sigma^2)^{t+1} \sim IG\left( \alpha_\nu + \frac{r}{2}, \beta_\nu + \frac{\bm \theta_r^T\bm \theta_r}{2\sigma^2}\right) \label{eq:sampling nu}
    \end{equation}
    \item The success rate $p_0^{(t)}$ is sampled from a Beta distribution as shown in \Eqref{eq:p derivation},
    \begin{equation}
        p_0^{(t+1)} \; \sim \; Beta(\alpha_p + h_Z, \beta_p + K - h_Z) \label{eq:sampling p}
    \end{equation}
    \item $G^{t+1}$ is sampled from a single Multinomial trial as shown in \Eqref{eq:G derivation},
    \begin{equation}
        G^{t+1}|\tau^{t+1} \; \sim \; Dirichlet(\alpha_0,\alpha_1,... \alpha_{j-1}, \alpha_{j}+1,\alpha_{j+1},...\alpha_{N-1}) \label{eq:sampling G}
    \end{equation}
    \item Finally, $\bm \theta$ is sampled from a Multivariate Gaussian distribution as,
    \begin{equation}
        p(\bm \theta^{t+1}|Z^{(+1},\nu^{t+1},{\sigma^2}^{t+1},\tau^{t+1}) \;\sim\; N(\mu,\sigma^2\Sigma) \label{eq:sampling theta}
    \end{equation}
    where $\Sigma^{-1} =L_{r,\tau}^TL_{r,\tau} + (\nu)^{-1}I_r$ and $\mu = \Sigma L_{r,\tau}^T\mathbf{Y}$.
\end{enumerate}

Overall, $N_{MC}$ iterations of sampling are conducted. The initial $\frac{N_{MC}}{4}$ iterations are discarded as burn-in samples to allow the Markov Chain Monte Carlo (MCMC) process to converge to a stationary distribution. The average value of the time delay indices obtained from the next $\frac{N_{MC}}{4}$ samples is regarded as the converged value of $\tau$. This converged value is then fixed and used for all subsequent iterations. Consequently, $\tau$ is not sampled during the final $\frac{N_{MC}}{2}$ iterations.
For the final step of equation discovery, the marginal posterior inclusion probability (PIP), denoted as $p(\bm Z | \mathbf{Y})$, is utilized. The PIP is computed for each of the $K$ candidate functions using the following expression:
\begin{equation}\label{eq:pip}
    p(Z_k = 1 | \mathbf{Y}) \approx \frac{1}{N_F} \sum\limits_{l=1}^{N_F} Z_k^l, \quad k = 1, 2, \dots, K,
\end{equation}
where $N_F$ denotes the number of post-burn-in samples and $Z_k^l$ represents the value of the $k$-th inclusion variable in the $l$-th sample. A PIP value exceeding 0.5 indicates that the corresponding candidate function was included in more than half of the posterior samples and is thus deemed likely to be part of the actual governing system. Any such function is included in the final model predicted by BayTiDe.
The expected value of the weight vector is computed as the mean of the sampled weight vectors across the posterior samples. Furthermore, all weights corresponding to functions with $PIP < 0.5$ are set to zero, thereby excluding those functions from the final model.

The initial latent vector is determined using a linear regression procedure, where weights with magnitudes less than 1\% of the maximum magnitude are set to zero. The initial noise variance is assigned as the mean squared error (MSE) of the model derived from this initial latent vector. The delay index $\tau$ is initialized arbitrarily to zero. 
The hyperparameters for the priors are specified as follows: $\alpha_p = 0.1$ and $\beta_p = 0.1$ for the Beta prior on $p_0$, $\alpha_\sigma = 0.0001$ and $\beta_\sigma = 0.0001$ for the inverse Gamma prior on the noise variance, $\alpha_\nu = 0.1$ and $\beta_\nu = 0.1$ for the inverse Gamma prior on the slab variance, and $\alpha_i = 1$ for $i = 0, 1, \dots, N-1$ for the Dirichlet prior on $G$.
The initial values of the remaining random variables are set as follows: $\nu = 0.5$, $p_0 = 0.1$, and $g_i = \frac{1}{N}$ for $i = 0, 1, \ldots, N-1$. 
The formal steps of the algorithm are presented in \hyperref[alg:cap]{Algorithm \ref{alg:cap}}.

\begin{algorithm}[ht]
\caption{The proposed BayTiDe algorithm for discovering delay differential equation from data}\label{alg:cap}
\begin{algorithmic}[1]
\Require $N_{MC}$, State Data $\mathbf X\in \mathbb{R}^{N\times m}$, hyper-parameters: $\alpha_p,\;\beta_p\;\alpha_\sigma,\;\beta_\sigma,\;\alpha_\nu,\;\beta_\nu,\;\{\alpha_i\}_{i=0}^{N-1}$
\Require Initial values of random variables: $p_0^{(0)},\; \nu^{(0)},\; \{g_i\}_{i=0}^{N-1}$, candidate functions, $N_{start},\;N_{end}$
\Note{All highly correlated candidate functions must be removed}
\State Choose random $\tau^{(0)}$ that is far outside the search range
\State Construct $\mathbf{L}_\tau^{(0)}$ from the sampled delay
\State Estimate initial latent vector $Z^{(0)}$ from linear regression using the library. Values less than 1\% of the largest weight are set to 0 and corresponding latent variables are considered inactive.
\State Estimate initial noise variance $\sigma^{(0)}$ from the MSE of the linear regression
\For{i = 1, 2, ..., $\text{N}_\text{MC}$}
    \State Update the delay variable $\tau^{(i)}$           \Comment{\Eqref{eq: sampling tau}}
    \State Update the library $\mathbf{L}_\tau^{(i)}$
    \State Update the latent vector $Z^{(i)}$               \Comment{\Eqref{eq:sampling Z}}
    \State Update the noise variance $\sigma^{(i)}$         \Comment{\Eqref{eq:sampling sigma}}
    \State Update the slab variance $\nu^{(i)}$             \Comment{\Eqref{eq:sampling nu}}
    \State Update the success rate $p_0^{(i)}$              \Comment{\Eqref{eq:sampling p}}
    \State Update the multinomial success rate $G^{(i)}$    \Comment{\Eqref{eq:sampling G}}
    \State Update the weight vector $\bm \theta^{(i)}$      \Comment{\Eqref{eq:sampling theta}}
\EndFor
\State Discard the burn-in samples  
\State Estimate the PIP for the latent vector               \Comment{\Eqref{eq:pip}}
\State Include the candidate functions with the required PIP values
\Ensure $Z$, $\tau$, $\bm \theta$
\end{algorithmic}
\end{algorithm}

\noindent
\textbf{Remark 1:}
In \Eqref{eq:zeta derivation}, the denominator comprises a summation of $N$ terms, each requiring the inversion of two $N \times N$ matrices to compute $\mu$ and $\Sigma$. This results in a computationally expensive procedure for sampling the delay index $\tau$. To mitigate this computational cost, we propose sampling the delay index within a smaller localized `window' rather than over all the $N$ time steps. 
\\\\
\noindent
\textbf{Remark 2:}
In \Eqref{eq:zeta derivation}, indices for which $\zeta_j \to 0$ are never sampled in subsequent iterations. To further optimize the sampling process, we propose `shrinking' the window dynamically to include only indices where $\zeta_j$ exceeds a predefined, arbitrarily small threshold, set to $10^{-100}$ in this case, after each iteration. The lower bound $N_{\text{lower}}$ is set to the smallest index that satisfies this threshold, while the upper bound $N_{\text{upper}}$ is set to the largest index meeting the condition. This modification significantly accelerates the convergence of the algorithm.
Additionally, if $\zeta_j = 1$ is encountered before the burn-in period concludes, the delay index $\tau$ is deemed to have converged and is no longer sampled in subsequent iterations. Mathematically, this behavior arises because $\zeta_j = 0$ and $\zeta_j = 1$ act as absorbing states. While these absorbing states may appear to violate the irreducibility condition, they are reached solely due to numerical overflow errors and do not formally belong to the Markov chain. These overflow errors occur as a result of the presence of $\omega_l$, which are ratios of probabilities and can lead to numerical instability.
\\\\
\noindent
\textbf{Remark 3:}
The starting index $N_{\text{start}}$ must be positioned sufficiently far from zero, as $\tau = 0$ represents an absorbing state. This arises because, as the delay index approaches zero, the correlation between the delayed and non-delayed terms converges to unity, resulting in inaccuracies within the algorithm.

\section{Numerical Studies} \label{sec:Numerical Studies}
In this section, we comprehensively evaluate the performance of the proposed BayTiDe algorithm using four canonical time-delay systems: (a) an exponential delay system, (b) the JC Sprott delay system, (c) the Mackey-Glass system, and (d) a linear coupled delay system. For each of these examples, we examine BayTiDe's ability to accurately discover the governing delay differential equations (DDEs) and compare predictions derived from the identified equations against the ground truth. Additionally, we explore BayTiDe's robustness to noise, its proficiency in estimating large time delays, and its capability to detect low-amplitude terms that contribute to the true governing equations. 
Lastly, we present comparative results demonstrating BayTiDe's superior performance over SINDy, even in cases where the exact delay term is provided to SINDy a priori.

The training data for all examples were synthesized using MATLAB's \texttt{dde23} solver. To simulate realistic scenarios, the generated data was corrupted with Gaussian white noise, with the noise variance set as fractions of the standard deviation of the training data. The first derivatives were approximated using a fifth-order finite-difference scheme. However, the sensitivity of this method to noise necessitated the use of a Butterworth filter to mitigate instabilities and enhance the reliability of derivative estimates. The filter parameters were selected empirically, as their optimal configuration depends on the specific characteristics of each system. 

\subsection{Exponential system}
\begin{figure}[H]
    \centering
    \begin{subfigure}[ht]{0.9\textwidth}
        \centering
        \includegraphics[width = \textwidth]{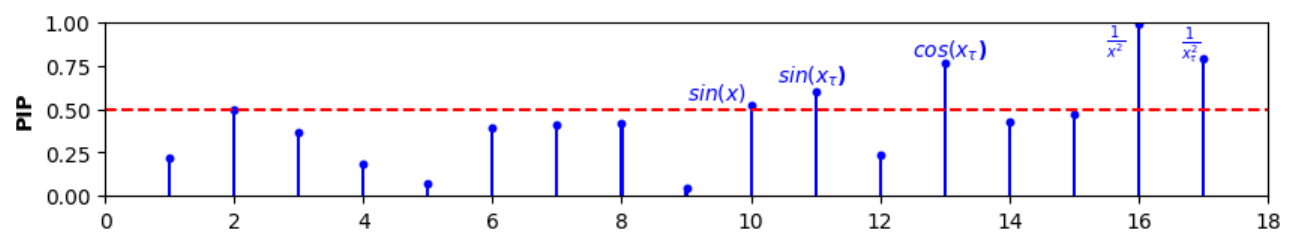}
        \caption{PIP when $\{\frac{1}{x},\; \frac{1}{x^2}\}$ are included.}
        \label{fig:exponential pip larger lib}
    \end{subfigure}
    \begin{subfigure}[ht]{0.9\textwidth}
        \centering
        \includegraphics[width = \textwidth]{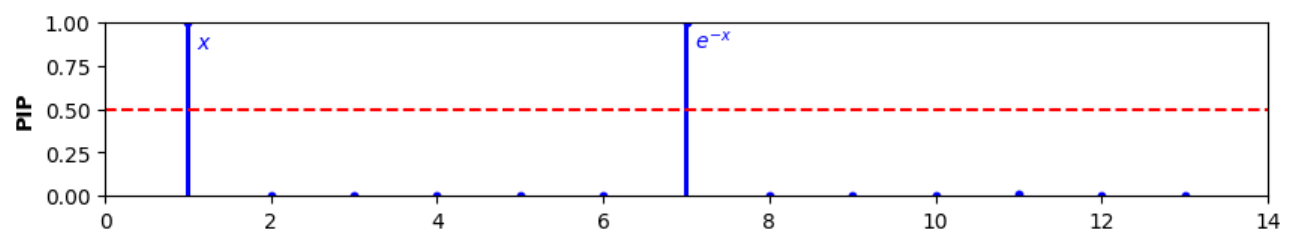}
        \caption{PIP when $\{\frac{1}{x},\; \frac{1}{x^2}\}$ are excluded.}
        \label{fig:exponential pip smaller lib}
    \end{subfigure}
    \caption{\textbf{PIP of the Exponential System corrupted with 15\% Gaussian white noise:} The feature library $\mathbf{L_\tau} \in \mathbb{R}^{n\times17}$ consists of the functions listed in \Eqref{eq:exponential function list}, applied combinatorially to the augmented data matrix. \autoref{fig:exponential pip larger lib} shows the PIP of each candidate function when the correlated functions are included. BayTiDe completely fails in identifying the ground truth. Another point to be noted is that the identified functions are expansions of the functions part of the real equation which represents a damped periodic system. \autoref{fig:exponential pip smaller lib} shows the PIP of each candidate function when the correlated functions are removed. The feature library now has the shape $\mathbf{L}^{n\times13}$. BayTiDe discovers the governing equation with a sure probability (PIP=1). The two figures highlight the dependency on the selection of candidate functions.}
    \label{fig:Exponential System results}
\end{figure}

\begin{figure}[H]
    \centering
    \begin{subfigure}[ht]{0.475\textwidth}
        \centering
        \includegraphics[width =\textwidth]{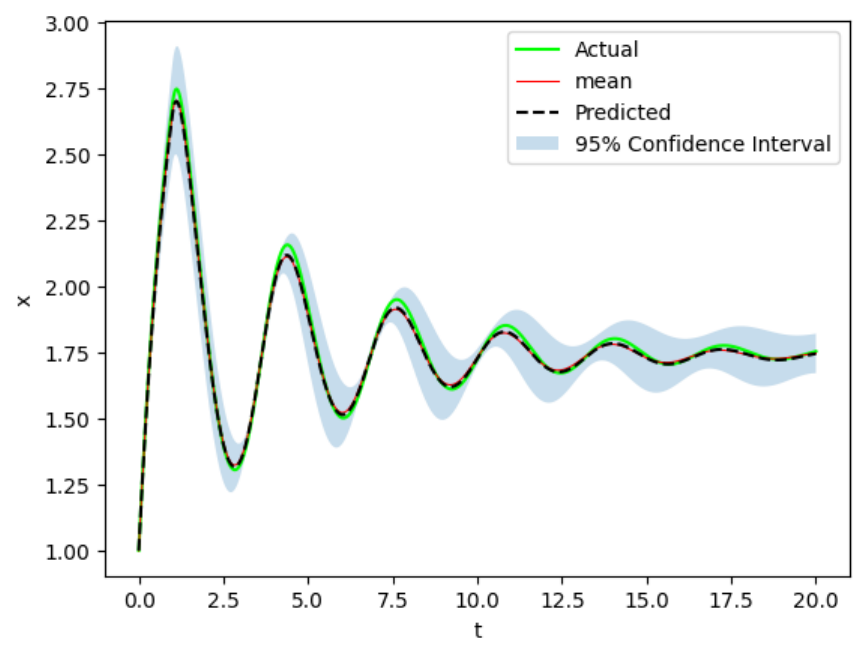}
        \caption{$x_0=1$.}
        \label{fig:exponential result diff initial}
    \end{subfigure}
    \begin{subfigure}[ht]{0.475\textwidth}
        \centering
        \includegraphics[width = \textwidth]{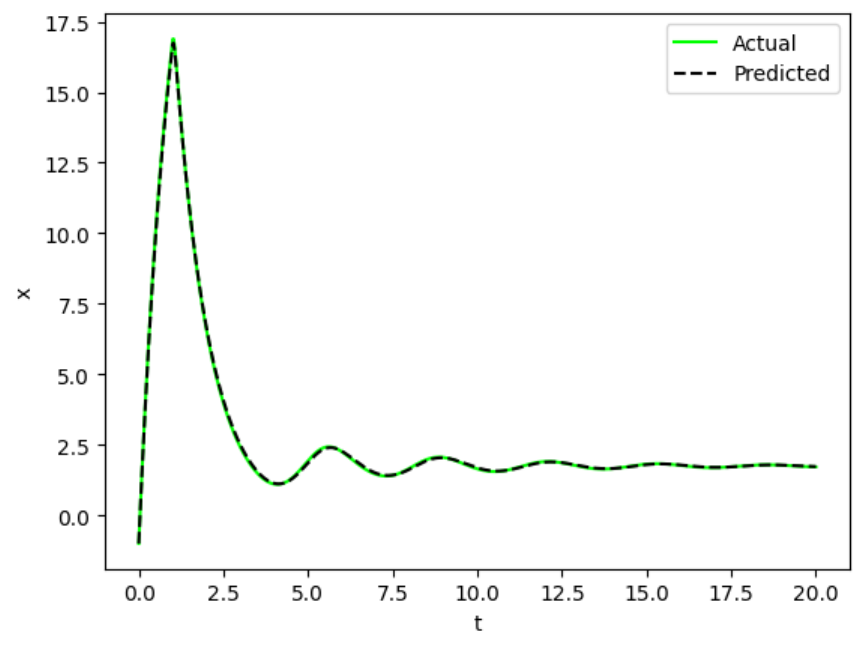}
        \caption{$x_0=-1$}
        \label{fig:exponential result training}
    \end{subfigure}
    \caption{\textbf{Uncertainty Plot and response comparison for the Exponential System:} Green: The real equation simulated with $\tau=1$. Dashed Line: The discovered equation simulated with discovered $\tau=0.99$ and the identified equation. \autoref{fig:exponential result diff initial} The shaded area is the 95\% CI of the predicted system, and represents the uncertainty associated with the predicted value at each time step. As such, it is observed to increase with time. The red line is the mean value of the predicted response after simulations using every sampled weight. \autoref{fig:exponential result training} Performance of the proposed approach against ground truth for a different initial condition that results in completely different system dynamics.}
    \label{fig:exponential plotss}
\end{figure}

As the first example, we consider an exponential time delay system to demonstrate BayTiDe's capacity to deal with non-linearities. The specific form of the system is given by,
\begin{equation}\label{eq:ex1}
    \dot{x} \;\; = \;\; 10\;e^{(-x_\tau)} \; - \; \;x,
\end{equation}
with $\tau=1$ and the initial value $x_0=1$. Synthetic data was generated by simulating the system for 20 seconds with $\Delta t = 0.01$ seconds. The generated data was corrupted with 15\% white Gaussian noise to emulate a realistic scenario. With this setup, the objective is to use the proposed BayTiDe algorithm to recover the governing equation in \Eqref{eq:ex1} from the noisy data. For this problem, the initial search window was set to $(N_{lower} = 20, N_{upper}=1000)$, and the library used for this test case was as follows:
\begin{equation} \label{eq:exponential function list}
    \mathbf{L} \;=\; \left( x\;\; x^2 \;\; xx_\tau \;\; e^{-x} \;\; e^x \;\; sin(x) \;\; cos(x) \;\; \frac{1}{x} \;\; \frac{1}{x^2} \right)
\end{equation}
where each function is independently applied to each of the state variables and the delay terms. This resulted in a library of size 18. Following Algorithm \ref{alg:cap}, we first check the correlation of the candidate function, which reveals a high correlation of 0.999. Noting the fact that $\frac{1}{x}$ and $\frac{1}{x^2}$ are included in the expansion of $e^{-x}$, and using human judgment, we retain $e^{-x}$ and proceed with the algorithm. The selected candidate functions are shown in \autoref{fig:exponential pip smaller lib}. We observe that the proposed algorithm has identified all the correct terms from the library. The discovered equation is as follows:
\begin{equation} \label{eq:predicted exponential}
    \dot{x} \; = \; \underset{\pm 0.2612}{9.78}\;e^{(-x_{\tau})} - \underset{\pm0.1317}{0.99}\;x, \quad \tau=0.99,
\end{equation}
which is very close to the original equation used shown in \Eqref{eq:ex1}. The coefficients are presented in the format $\underset{\pm sd}{mean}$. This indicates that BayTiDe identifies the delay term, the candidate functions, and the associated parameters with a high order of accuracy, even when the data is corrupted with 15\% noise. 
The performance of the proposed BayTiDe is further evident by the fact that the predicted response using the identified equation for an entirely new initial condition matches exactly with the ground truth response obtained using the original equation (see \autoref{fig:exponential result diff initial}). 
We carried out an ablation study to illustrate the effect of candidate function filtering based on correlation. To that end, we reemployed the algorithm to the same dataset but without the filtering step. The identified candidate functions are shown in Fig. \ref{fig:exponential pip larger lib}. We observe that without filtering, the proposed approach fails to capture the correct candidate functions. This illustrates the necessity of the filtering step and the need for selecting the correct candidate functions in the library.

\subsection{JC Sprott System}
As the second example, we analyze the JC Sprott system, which is governed by the delay differential equation (DDE) given as:
\begin{equation}\label{eq:jc sprott}
    \dot{x} = \sin(x_\tau).
\end{equation}
This system is a widely studied example of DDEs featuring sinusoidal nonlinearity, known for its ability to exhibit both periodic and chaotic behavior \cite{JCSPROTT}. 
For generating the training data, we set the delay parameter $\tau = 3$, the initial condition $x_0 = 0.1$, and the time-step $\Delta t = 0.05$. Using these parameters, data was generated for a simulation duration of 100 seconds, following the same procedure as employed in the first example. The goal of this experiment is to identify the governing DDE from the noisy data. For this purpose, the initial search window was defined with bounds $N_{lower} = 20$ and $N_{upper} = 1000$. 
The library of candidate functions used for this example is identical to the one used in the previous example:
\begin{equation} \label{eq:jc sprott functions}
    \mathbf{L} = \left( x, \; x^2, \; xx_\tau, \; e^{-x}, \; \sin(x), \; \cos(x), \; \frac{1}{x}, \; \frac{1}{x^2} \right).
\end{equation}
The BayTiDe framework, as outlined in Algorithm~\ref{alg:cap}, was applied to discover the governing equation. 

Figure~\ref{fig:jc sprott pip} shows the PIP for each candidate function. The correct term, $\sin(x_\tau)$, is identified with a PIP close to 1. All other candidate functions exhibit PIPs well below the threshold value of 0.5, except for $e^{-x}$, which has a PIP of 0.48. Nonetheless, the PIP of $e^{-x}$ does not exceed the threshold, and thus it does not appear in the final identified equation. Additionally, BayTiDe successfully identifies the delay parameter $\tau = 3$ with high accuracy.
The equation identified using BayTiDe is:
\begin{equation}
    \dot{x} = \underset{\pm 0.029}{0.984} \sin(x_\tau), \quad \tau = 3,
\end{equation}
which closely approximates the original governing equation. We further compare the responses obtained using the discovered equation with those obtained using the original equation. We illustrate two cases, one with the same initial condition as the training data (Fig.~\ref{fig:jc sprott tau 3 training}) and one with a different initial condition (Fig.~\ref{fig:jc sprott tau 3 diff initial}). 
In both cases, the response predicted by the identified equation matches the ground truth obtained from the original equation with high fidelity. 
This again demonstrates the robustness and accuracy of the BayTiDe framework in recovering the governing equations of complex time-delay systems, even under noisy conditions.
\begin{figure}[ht]
     \centering
     \begin{subfigure}[b]{\textwidth}
         \centering
         \includegraphics[width=\textwidth]{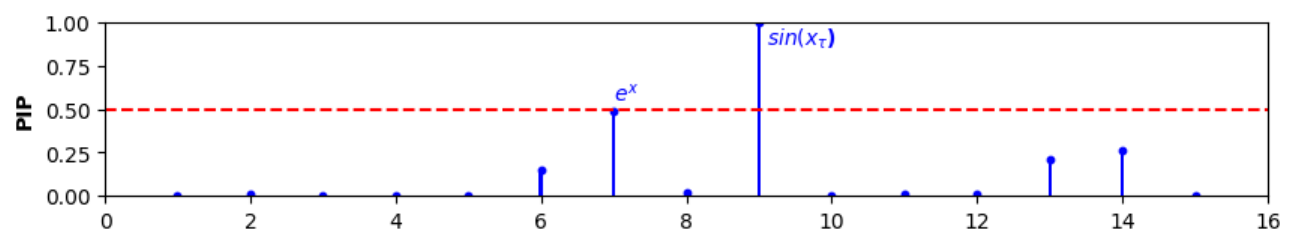}
         \caption{PIP of the candidate functions}
         \label{fig:jc sprott pip}
     \end{subfigure}
     \begin{subfigure}[b]{0.475\textwidth}
         \centering
         \includegraphics[width=\textwidth]{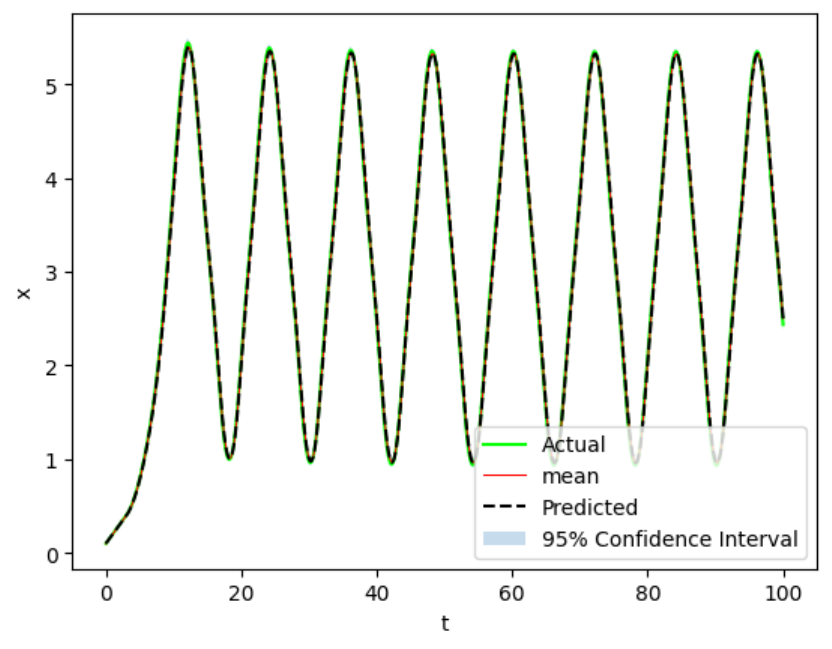}
         \caption{$\tau=3, \; x_0=0.1$.}
         \label{fig:jc sprott tau 3 training}
     \end{subfigure}
     \hspace{0.0025\textwidth}
      \begin{subfigure}[b]{0.475\textwidth}
         \centering
         \includegraphics[width=\textwidth]{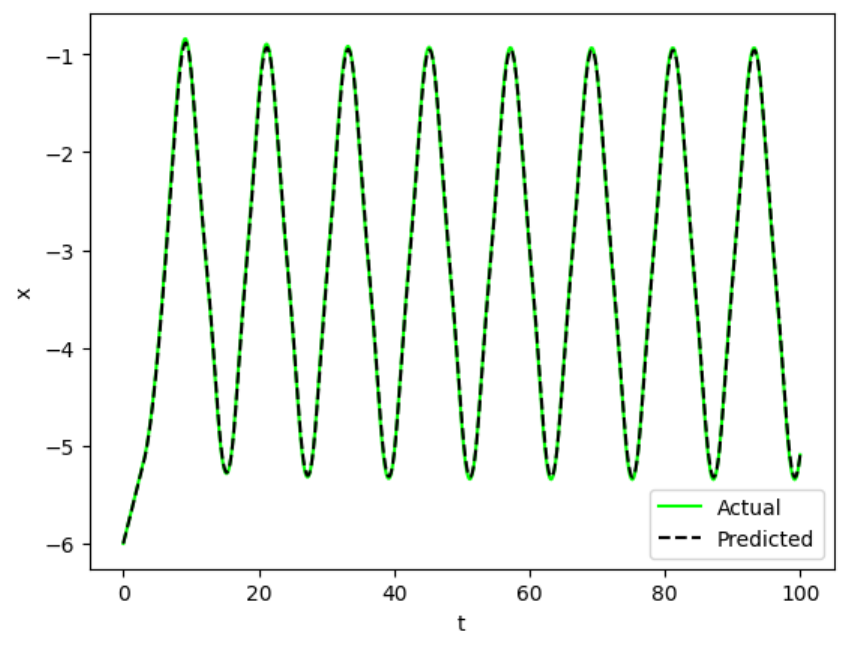}
         \caption{$\tau=3, \;x_0=-6$}
         \label{fig:jc sprott tau 3 diff initial}
     \end{subfigure}
     \caption{\textbf{Performance of the proposed approach for JC Sprott System corrupted with 15\% Gaussian white noise:} The feature library $\mathbf{L_\tau}\in\mathbb{R}^{n\times15}$ consists of the functions listed in \Eqref{eq:jc sprott functions}. \autoref{fig:jc sprott pip} shows the PIP of the candidate functions. The function $\sin(x_\tau)$ is identified correctly with a PIP=1.  $e^{x_\tau}$ 
     has a PIP=0.48; however, it does not appear in the identified equation as it is below the threshold. \autoref{fig:jc sprott tau 3 training} compares the response obtained using the identified equation with ground truth. The 95\% CI (the shaded region) is overlapped by the plots, indicating high confidence in the identified model. \autoref{fig:jc sprott tau 3 diff initial} compares the equation to the actual system for a different initial condition.}
     \label{fig: jc sprott results}
\end{figure}

\subsection{Mackey-Glass System}\label{sec:Chaotic systems}
In this example, we consider the well-known Mackey-Glass time delayed system,
\begin{equation}
    \dot{x} = -0.1\;x \; + \; \frac{0.2\;x_\tau}{1+x_\tau^{10}} \label{eq:mackey glass equation}
\end{equation}
This is an extremely challenging problem as it exhibits hyper-chaotic behavior for time delays $>$ 32 \cite{mackey-glass}. We consider $\tau=40$ to demonstrate BayTiDe's capacity to identify large time delays in hyperchaotic systems. For generating training data, we considered $\Delta t = 0.2$ seconds and the initial condition $x_0=0.1$, and used the same procedure as the previous examples to generate data until 1000 secs. The generated data was further corrupted with 15\% white Gaussian noise to emulate a realistic scenario. With this setup, the objective is to use the training data to identify the governing DDE using the proposed BayTiDe framework.
To that end, we select the following library:
\begin{equation} \label{eq:mackey glass functions}
    \mathbf{L} \;\;\;=\;\;\; \left[ x \;\; x^2 \;\; sin(x) \;\; cos(x) \;\; \frac{x}{1+x^{10}} \;\; \frac{x}{1+x^4} \;\; \frac{1}{x} \;\; \frac{1}{x^2}\right],
\end{equation}
where similar to the previous examples, each function is applied to both $x$ and $x_\tau$. We employ Algorithm \ref{alg:cap} to discover the governing equation. The PIPs for all the candidate functions are shown in Fig. \ref{fig:mackey glass pip}. We observe that one additional candidate function is identified. However, the associated parameter value is $4\times 10^{-4}$, and hence, this term is ultimately ignored. The time-delay $\tau=40$ is exactly identified. Overall, the final identified equation is
\begin{equation}
    \dot{x} \; =\; -0.0998\;x \; +\; 0.1988\; \frac{x_{\tau}}{1+x_{\tau}^{10}} \; + \; \underbrace {0.0004 \; \frac{1}{x_{\tau}}}_{ignored} = \dot{x} \; =\; \underset{\pm0.0043}{-0.0998}\;x \; +\;\underset{\pm0.0063}{0.1988}\; \frac{x_{\tau}}{1+x_{\tau}^{10}}, \quad \tau=40.
\end{equation}
While the identified equation is quite close to the original equation, the hyper-chaotic behavior necessitates further investigation. To that end, the response obtained using the identified equation is compared to the response obtained using the original DDE. The corresponding results are shown in Fig. \ref{fig:mackey glass equation plot}. We observe that the response obtained using the identified equation starts deviating beyond 800 secs. The mean response obtained by sampling the weights of the identified equation and then simulating, on the other hand, deviates beyond 500 seconds. Nonetheless, the uncertainty bound encompasses the ground truth response. It is worth noting that as the response obtained using the identified equation starts deviating from the ground truth response, the associated predictive uncertainty also increases. This indicates that predictive uncertainty is a useful metric for identifying the deviation of the predictive response. Overall, this clearly demonstrates the challenge associated with discovering DDE in hyperchaotic systems and the importance of an Bayesian approach.

As an additional case study, we reemploy BayTiDe to discover the governing DDE for case where the training data is corrupted by 10\% noise. The identified DDE for this case is\begin{equation}
    \dot{x} \; =\; -\underset{\pm0.0016}{0.1003}\;x \; +\; \underset{\pm0.0032}{0.1999}\; \frac{x}{1+x^{10}},  \quad \tau=40.
\end{equation} 
As expected, the equation identified is more accurate.
The response obtained using the identified equation and the ground truth response is shown in 
\autoref{fig:mackey glass less noise equation plot}. The response obtained by solving the identified equation matches exactly with the ground truth response. The mean response, on the other hand, deviates slightly towards the end. Additionally, we observe a reduction in uncertainty. Overall, this example reinforces the strength of the proposed BayTiDe in handling longer time-delay and hyperchaotic systems with a reasonable accuracy.

\begin{figure}[ht]
    \centering
    \begin{subfigure}[h]{\linewidth}
        \centering
        \includegraphics[width=\linewidth]{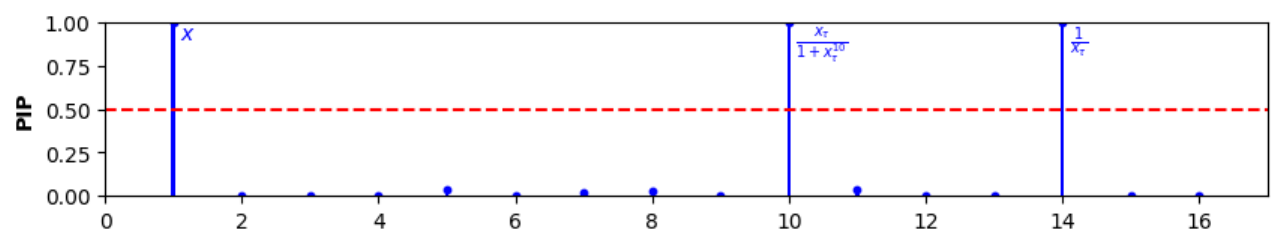}
        \caption{The final PIP values of the chosen candidate functions in both runs.}
        \label{fig:mackey glass pip}
    \end{subfigure}
    \begin{subfigure}[h]{0.49\linewidth}
        \centering
        \includegraphics[width = \linewidth]{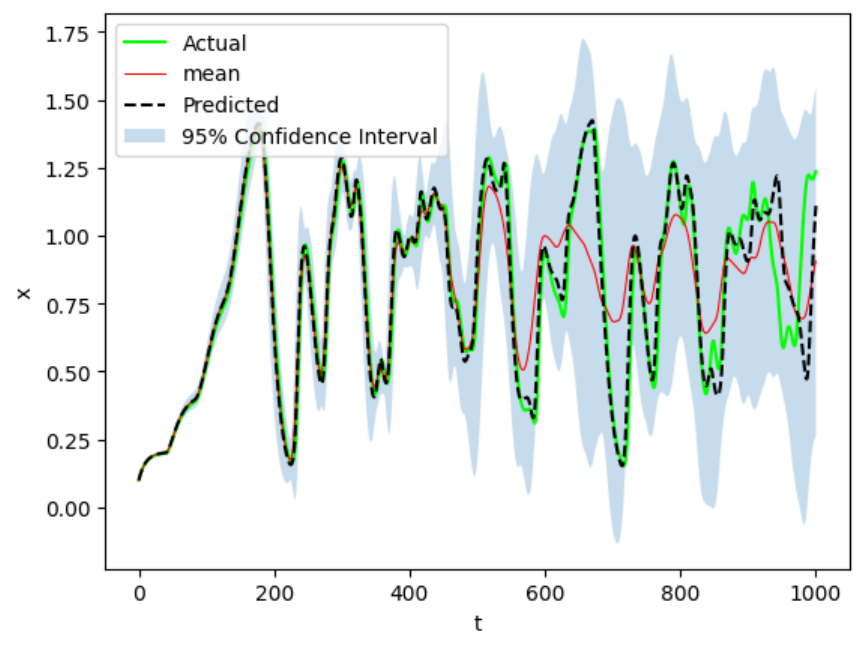}
        \caption{$x_0=0.1$, 15\% Noise.  }
        \label{fig:mackey glass equation plot}
    \end{subfigure}
    \ContinuedFloat
    \begin{subfigure}[h]{0.49\linewidth}
        \centering
        \includegraphics[width = \linewidth]{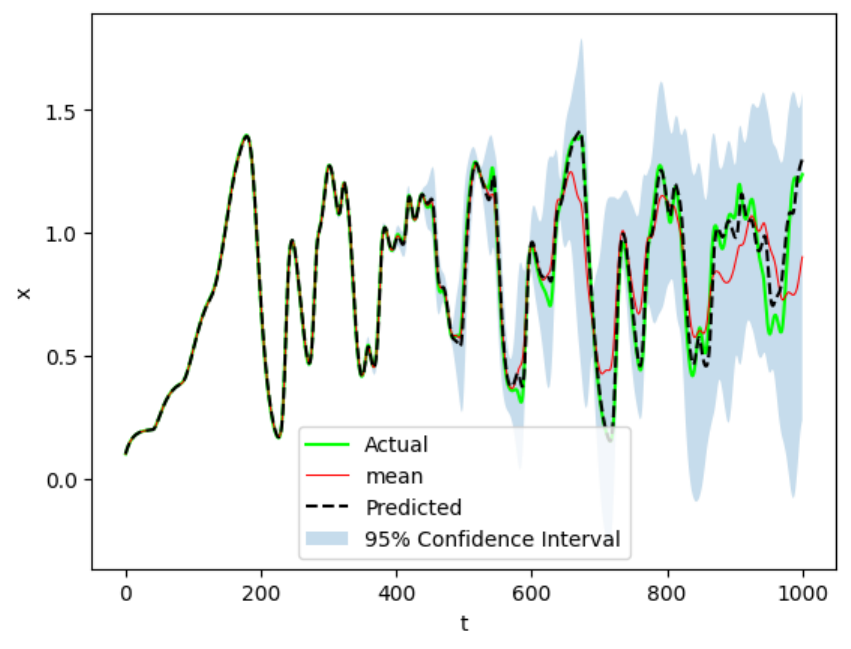}
        \caption{$x_0=0.1$, 10\% Noise.}
        \label{fig:mackey glass less noise equation plot}
    \end{subfigure}
    \caption{\textbf{Performance of BayTiDe for Mackey-Glass System:} 
    \autoref{fig:mackey glass pip} illustrates the PIP for the system corrupted with both levels of noise. The feature library $\mathbf{L_\tau}\in\mathbb{R}^{n\times16}$ consists of functions listed in \Eqref{eq:mackey glass functions}. For both cases, the PIP of the candidate functions were identical, with a change only in the accuracy of the prediction and a reduction in the uncertainty associated, as shown in \autoref{fig:mackey glass equation plot} and \autoref{fig:mackey glass less noise equation plot}. Shaded area: the predictive uncertainty associated with each time step. Red line: Indicates the average value of the state variables after simulation using the sampled weights. Dotted Line: response of the equation identified by BayTiDe.
    }
    \label{fig:mackey glass results}
\end{figure}

\subsection{2 Degree of Freedom Linear System}
\begin{figure}[ht]
    \centering
     \begin{subfigure}[h]{\linewidth}
        \centering
        \includegraphics[width=\linewidth]{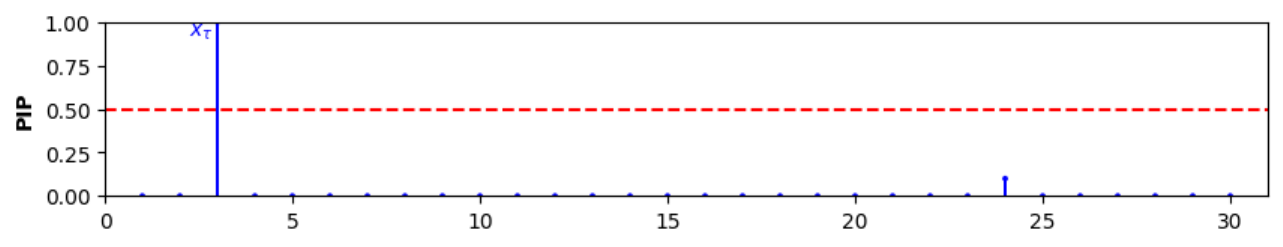}
        \caption{The final PIP values of the chosen candidate functions in predicting \Eqref{eq: linear system xdot}.}
        \label{fig:linear system pip x}
    \end{subfigure}
     \begin{subfigure}[h]{\linewidth}
        \centering
        \includegraphics[width=\linewidth]{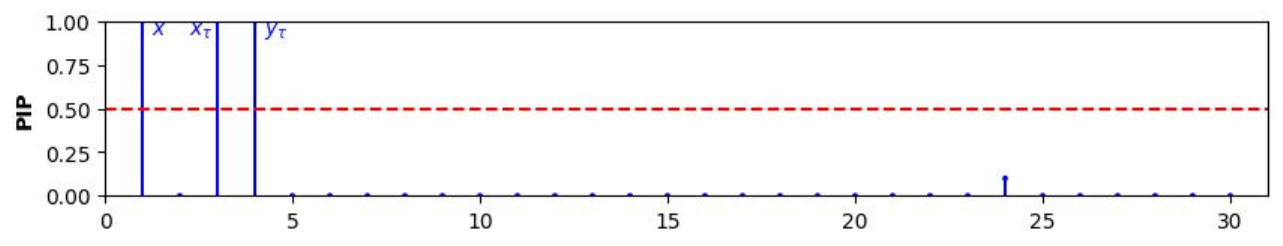}
        \caption{The final PIP values of the chosen candidate functions in predicting \Eqref{eq: linear system ydot}.}
        \label{fig:linear system pip y}
    \end{subfigure}
    \begin{subfigure}[b]{0.475\textwidth}
         \centering
         \includegraphics[width=\textwidth]{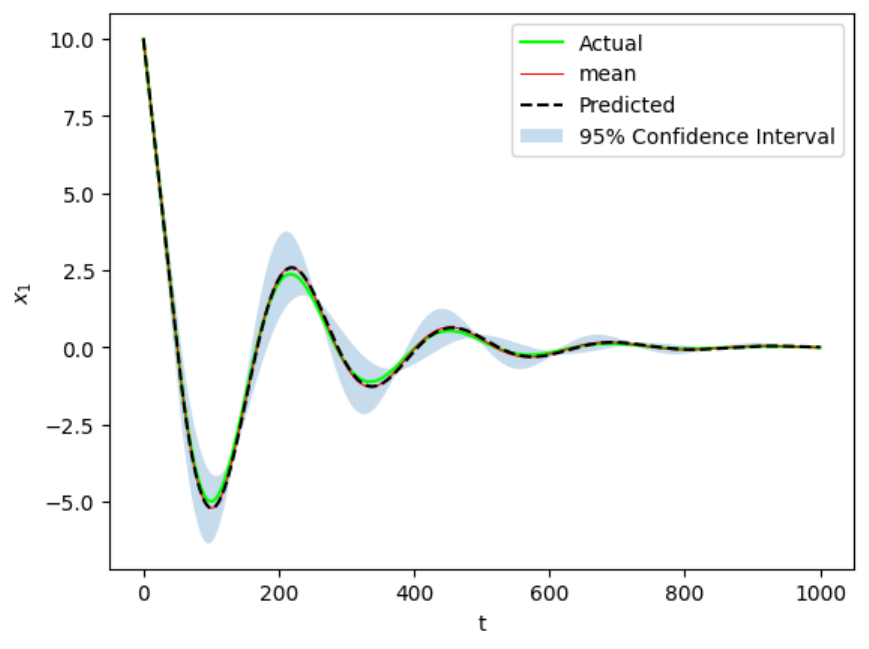}
         \caption{$\{x_0,y_0\}=\{10,6\}$.}
         \label{fig:linear coupled system x}
     \end{subfigure}
     \hspace{0.0025\textwidth}
      \begin{subfigure}[b]{0.475\textwidth}
         \centering
         \includegraphics[width=\textwidth]{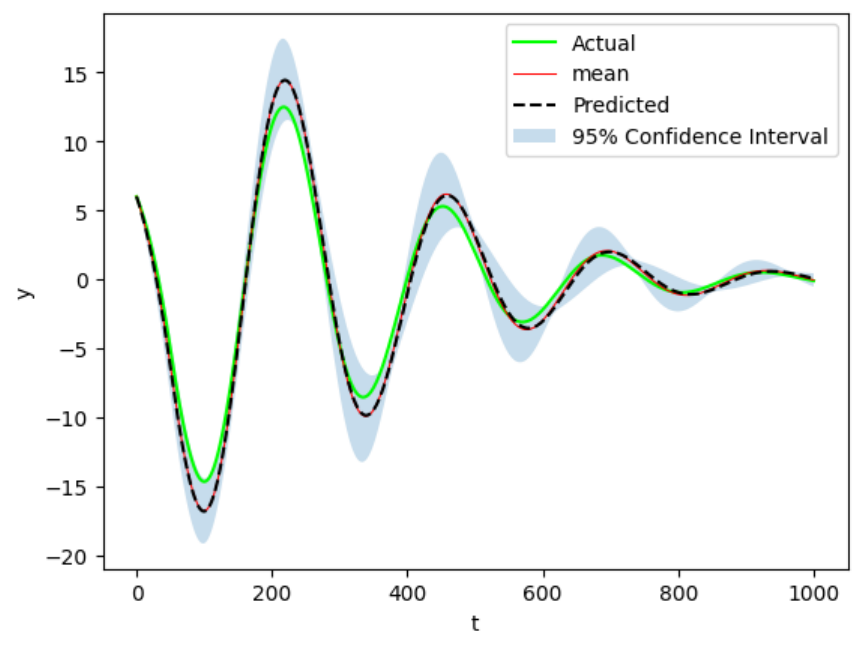}
         \caption{$\{x_0,y_0\}=\{10,6\}$.}
         \label{fig:linear coupled system y}
     \end{subfigure}
     \caption{\textbf{Performance of BayTiDe for the Linear 2-DOF coupled system:} \autoref{fig:linear system pip x} represents the PIP of the candidate functions when identifying the DDE for the first DOF. BayTiDe identifies the ground truth with complete confidence (PIP=1). \autoref{fig:linear system pip y} represents the PIP of the functions when identifying the DDE for the second DOF. Once again, the functions are identified with complete confidence. \autoref{fig:linear coupled system x} and \autoref{fig:linear coupled system y} compare the predicted system (dotted line) with the ground truth (green line). The red line represents average response of the simulated weight samples, and the shaded area represents the 95\% CI of the predictive uncertainty. The real response appears to be at the edge of the confidence interval possibly because time delay used was the average of the two identified.} 
     \label{fig:linear coupled system results}
\end{figure}

As the last example, we consider a coupled system of DDEs,
\begin{align}
    \dot{x} \; &= \; -x_\tau  \label{eq: linear system xdot} \\
    \dot{y} \; &= \; x \; - \; x_\tau \; - \; y_\tau \label{eq: linear system ydot}
\end{align}
The purpose of this example is to demonstrate BayTiDes ability to discover the governing equations of coupled systems of DDEs. 
For generating synthetic data, we consider $\tau = 1$,  initial conditions $\{x_0,y_0\} = {10,6}$, and $\Delta t = 0.01$. Training data is generated following the same procedure as before by simulating the system for 20 secs.
The data generated was corrupted with 20\% Gaussian white noise. With this setup, we employ Algorithm \ref{alg:cap} to identify the underlying coupled DDE from the noisy data. The same candidate functions as the exponential system were used for this example. The functions $\{\frac{1}{x},\; \frac{1}\}$ were removed from the library as the state variables were observed to change sign within the measurement period. The PIP for the candidate functions are shown in Figs. \ref{fig:linear coupled system results} (a) and \ref{fig:linear coupled system results} (b). Note that the algorithm was employed independently for each degree of freedom.  Naturally, this resulted in two slightly different values of identified delay. 
Taking the average of the two, the resulting equation identified using BayTiDe is as follows: 
\begin{align}
    \dot{x} \; &= \; -\underset{\pm0.115}{1.000}\;x_\tau \\
    \dot{y} \; &= \; \underset{\pm0.23}{1.018}\;x \; - \underset{\pm0.227}{1.172}\; x_\tau \; -\underset{\pm0.096}{0.952} \; y_\tau, \quad \tau=1.02
\end{align}
The predictions are shown in \autoref{fig:linear coupled system results} (c and d).
We observe that the responses obtained using the predicted equation match almost exactly with the ground truth response. This indicates that the proposed BayTiDe performs equally well for coupled DDEs.

\subsection{Generalizing for unseen poles}
\begin{figure}[ht]
    \centering
    \includegraphics[width=0.95\linewidth]{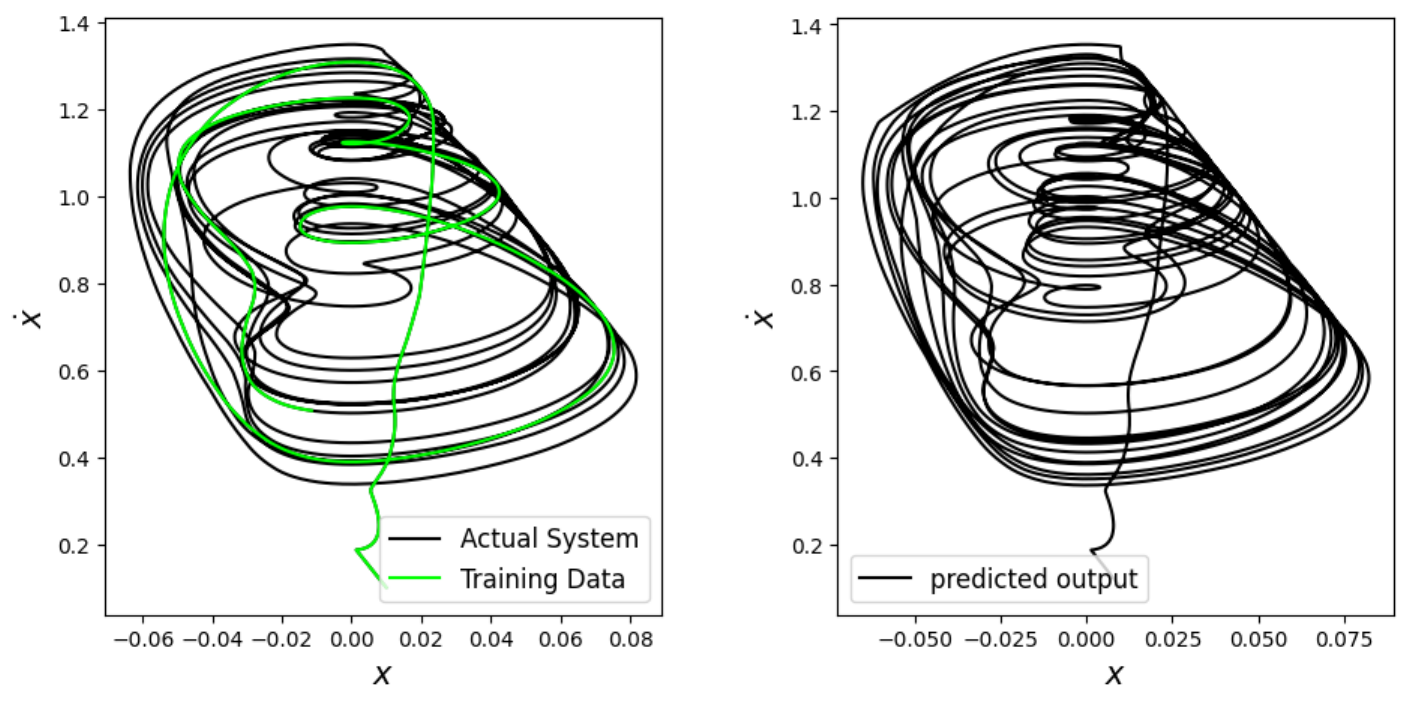}
    \caption{Phase diagram of the Mackey Glass system for $\tau=20$. \textbf{Left:} The actual system. Green line represents the first 2000 time steps which were input to the system. \textbf{Right:} The predicted system. All the modes of the actual system were predicted without being input to the system.}
    \label{fig:poles plot}
\end{figure}

Before concluding this section, we present two additional case studies to further highlight the strengths of the proposed BayTiDe approach. The first case study investigates the ability of BayTiDe to generalize and predict unseen poles in a dynamical system. For this purpose, we revisit the Mackey-Glass system previously discussed. In this case, we consider a time delay $\tau = 20$ and generate data for up to 10,000 seconds, following the same setup as described earlier. The data corresponding to the first 1,000 seconds is used as training data to identify the governing delay differential equation. A phase diagram for the system is presented in Fig.~\ref{fig:poles plot}(a), where the training data is highlighted in green. Notably, the training data only represents a subset of the system's poles, leaving several poles unobserved during the training phase.
The objective of this study is to evaluate whether the proposed BayTiDe algorithm can accurately capture the complete set of poles, including those not present in the training data. To address this, BayTiDe is employed to infer the governing DDE by following the procedure outlined in Algorithm~\ref{alg:cap}. The resulting equation, discovered using BayTiDe, is subsequently used to generate the phase diagram of the system, shown in Fig.~\ref{fig:poles plot}(b). 
The results demonstrate that the response obtained using the BayTiDe-derived equation successfully captures all the poles of the system, including those absent in the training data. This remarkable generalization capability underscores the strength of BayTiDe in accurately learning and reproducing the dynamics of time-delay systems, even in scenarios where the training data only partially represents the system's behavior. The ability to extrapolate system dynamics to unobserved regions of the phase space highlights the robustness and predictive power of the proposed approach.

\subsection{Comparison with SINDy}
As the final case study, we evaluate the performance of the proposed BayTiDe framework in comparison with the well-established SINDy approach \cite{brunton2016discovering}. While SINDy has been extensively used for discovering governing equations in dynamical systems, it is not inherently designed to handle time-delay systems. To facilitate a fair comparison, we manually provide SINDy with the exact time delay for the system under consideration, thereby allowing it to bypass the challenge of time-delay estimation. In contrast, the BayTiDe framework is tasked with simultaneously identifying both the governing equations and the unknown time delay.
The results of this comparison are summarized in Table~\ref{tab:error table}, where the mean squared error (MSE) of the reconstructed equation (parameters) is reported for varying noise levels. 
The results clearly demonstrate the superiority of the proposed BayTiDe framework. For all noise levels, BayTiDe achieves an MSE that is at least an order of magnitude lower than that obtained by SINDy. This finding is particularly noteworthy given the considerable advantage afforded to SINDy through the provision of the exact delay term, which BayTiDe has to infer from noisy data. 
This exceptional performance of BayTiDe can be attributed to its Bayesian framework, enabling it to handle noisy and sparse data more effectively. In contrast, SINDy relies on sparse regression techniques that are more sensitive to noise, especially when dealing with high-dimensional libraries of candidate functions.
This comparison underscores the robustness and accuracy of BayTiDe in discovering the underlying dynamics of time-delay systems, even in the presence of significant noise. The ability of BayTiDe to simultaneously infer unknown delays and identify governing equations from noisy data further highlights its potential as a powerful tool for analyzing complex time-delay systems.

\begin{table}[ht]
    \centering
    \caption{Comparison between SINDy and BayTiDe: A point to note is that SINDy was given the exact value of time delay before executing the equation discovery. As such, BayTiDe was given a significant handicap in these comparisions yet shows more accurate results.}
    \begin{tabularx}{\textwidth}{l|llY|llY|llY}
    \hline
        \textbf{Noise}(\%) & \multicolumn{9}{c}{\textbf{Parameter Estimation Error $e_\theta$}} \\ \cline{2-10}
         && \multicolumn{2}{l|}{JC Sprott} & & \multicolumn{2}{l|}{Exponential} & & \multicolumn{2}{l}{Mackey-Glass} \\ \cline{2-10}
            & SINDy                & BayTiDe & & SINDy & BayTiDe &  & SINDy & BayTiDe &  \\ \hline
         5  & $4\times10^{-6}$     &   $1.6\times10^{-7}$   & &   $1\times10^{-4}$    &  $3.7\times10^{-5}$ & & $4\times10^{-4}$   & $2\times10^{-6}$ & \\
         10 & $2.5\times10^{-5}$   &   $1.6\times10^{-7}$   & &   $4.3\times10^{-4}$ &  $2\times10^{-4}$    & & $4\times10^{-4}$   & $2\times10^{-6}$ & \\
         15 & $6.4\times10^{-5}$   &   $1.6\times10^{-7}$   & &   $0.611$            &  $7.1\times10^{-4}$  & & $9\times10^{-4}$   & $5.8\times10^{-5}$ & \\
         20 & $1.7\times10^{-2}$   &   $2\times10^{-3}  $   & &   $0.6$              &  $1\times10^{-3}$    & & $1.9\times10^{-3}$ & $6.9\times10^{-4}$ & \\ \hline
    \end{tabularx}
    \label{tab:error table}
\end{table}
\section{Conclusion} \label{sec:conslusion}
In this work, we introduced BayTiDe, a novel Bayesian framework for discovering governing delay differential equations (DDEs) from noisy and sparse data. The proposed approach integrates Bayesian inference with an efficient search strategy over delay indices, enabling accurate identification of both the functional form and the time delays inherent in time-delay systems. To exploit the fact that the physics model is generally parsimonious, BayTiDe exploits sparsity-promoting spike-and-slab prior in conjunction with Gibbs sampling. 
Unlike existing methods such as SINDy, BayTiDe is explicitly designed to handle time-delay dynamics without requiring prior knowledge of the delays, making it a versatile and powerful tool for modeling complex systems. The key features of the proposed BayTiDe framework include its robustness to noise, its capability to handle highly nonlinear and hyperchaotic systems, and its ability to seamlessly handle large time delays.

The efficacy of BayTiDe was demonstrated using several benchmark examples, including the exponential delay system, JC Sprott system, Mackey-Glass system, and a linear coupled delay system. Through these case studies, we showcased BayTiDe’s robustness to noise, its ability to accurately identify large delays, and its capability to uncover low-amplitude terms contributing to the governing equations. Notably, BayTiDe consistently outperformed SINDy, even when the exact delay terms were provided to SINDy a priori, highlighting the superior accuracy and reliability of the proposed approach.
The robustness of BayTiDe was further illustrated by its ability to generalize system dynamics to unobserved regions of the phase space. For instance, in the Mackey-Glass system, BayTiDe successfully identified all poles of the system, including those absent from the training data, underscoring its remarkable extrapolation capabilities. Moreover, the algorithm was shown to converge efficiently by dynamically shrinking the search window based on posterior probabilities, thereby accelerating computation without compromising accuracy. Overall, BayTiDe offers a comprehensive framework for discovering time-delay dynamics with minimal prior knowledge. 

Despite the excellent performance of BayTiDe, there are a few limitations associated with the framework. The selection of candidate functions plays a critical role in its performance; the presence of highly correlated candidate functions can lead to misidentification or even complete failure of the algorithm. Although this has been addressed to a degree by using candidate function filtering based on correlation and human intervention, a more elegant approach is warranted. Additionally, BayTiDe assumes that the input data is measured over uniform time intervals, with the time step sufficiently smaller than the time delay to be identified. The current framework is also limited to identifying a single constant delay, restricting its applicability to systems with more complex delay structures. Future research could extend BayTiDe to handle multiple constant delays or time-dependent delays, thereby broadening its applicability. Furthermore, while the framework performs well in discovering system dynamics, the accuracy of the identified weights for chaotic systems can still be improved, as the predictive power of the discovered equations may degrade over time. Addressing these limitations would enhance BayTiDe’s robustness and extend its utility in modeling a wider class of time-delay systems.

\section*{Acknowledgement}
 SC acknowledges the financial support received from Anusandhan National Research Foundation (ANRF) via grant no. CRG/2023/007667 and from the Ministry of Port, Shipping, and Waterways via letter no. ST-14011/74/MT (356529).


\begin{thebibliography}{10}

\bibitem{neurosciencetimedelay}
Mukeshwar Dhamala, Viktor~K. Jirsa, and Mingzhou Ding.
\newblock Enhancement of neural synchrony by time delay.
\newblock {\em Phys. Rev. Lett.}, 92:074104, Feb 2004.

\bibitem{timedelayepidemiology}
Oluwatosin Babasola, Oshinubi Kayode, Olumuyiwa~James Peter, Faithful~Chiagoziem Onwuegbuche, and Festus~Abiodun Oguntolu.
\newblock Time-delayed modelling of the covid-19 dynamics with a convex incidence rate.
\newblock {\em Informatics in Medicine Unlocked}, 35:101124, 2022.

\bibitem{economicstimedelay}
A~Krawiec and M~Szydlowski.
\newblock The kaldor‐kalecki business cycle model.
\newblock {\em Ann. Oper. Res.}, 89:89--100, 1999.

\bibitem{engineeringtimedelay}
Dawei~Zhang Zifan~Gao, Tao~Wu and Shuqian Zhu.
\newblock Network-based gain-scheduled control for preview path tracking of autonomous electric vehicles subject to communication delays.
\newblock {\em International Journal of Systems Science}, 53(12):2549--2565, 2022.

\bibitem{schmidt2009distilling}
Michael Schmidt and Hod Lipson.
\newblock Distilling free-form natural laws from experimental data.
\newblock {\em science}, 324(5923):81--85, 2009.

\bibitem{Karniadakis2021}
George~Em Karniadakis, Ioannis~G. Kevrekidis, Lu~Lu, Paris Perdikaris, Sifan Wang, and Liu Yang.
\newblock Physics-informed machine learning.
\newblock {\em Nature Reviews Physics}, 3(6):422--440, Jun 2021.

\bibitem{1100705}
H.~Akaike.
\newblock A new look at the statistical model identification.
\newblock {\em IEEE Transactions on Automatic Control}, 19(6):716--723, 1974.

\bibitem{schwarz1978estimating}
Gideon Schwarz.
\newblock Estimating the dimension of a model.
\newblock {\em The annals of statistics}, pages 461--464, 1978.

\bibitem{symbolic_regression}
Josh Bongard and Hod Lipson.
\newblock Automated reverse engineering of nonlinear dynamical systems.
\newblock {\em Proceedings of the National Academy of Sciences}, 104(24):9943--9948, 2007.

\bibitem{genetic-programming}
Michael Schmidt and Hod Lipson.
\newblock Distilling free-form natural laws from experimental data.
\newblock {\em Science}, 324(5923):81--85, 2009.

\bibitem{SINDy}
Steven~L. Brunton, Joshua~L. Proctor, and J.~Nathan Kutz.
\newblock Discovering governing equations from data by sparse identification of nonlinear dynamical systems.
\newblock {\em Proceedings of the National Academy of Sciences}, 113(15):3932--3937, 3 2016.

\bibitem{boninsegna2018sparse}
Lorenzo Boninsegna, Feliks N{\"u}ske, and Cecilia Clementi.
\newblock Sparse learning of stochastic dynamical equations.
\newblock {\em The Journal of chemical physics}, 148(24):241723, 2018.

\bibitem{Mangan201652}
Niall~M. Mangan, Steven~L. Brunton, Joshua~L. Proctor, and J.~Nathan Kutz.
\newblock Inferring biological networks by sparse identification of nonlinear dynamics.
\newblock {\em IEEE Transactions on Molecular, Biological, and Multi-Scale Communications}, 2(1):52 – 63, 2016.
\newblock Cited by: 272; All Open Access, Green Open Access.

\bibitem{bridge_sindy}
Shanwu Li, Eurika Kaiser, Shujin Laima, Hui Li, Steven~L. Brunton, and J.~Nathan Kutz.
\newblock Discovering time-varying aerodynamics of a prototype bridge by sparse identification of nonlinear dynamical systems.
\newblock {\em Physical Review E}, 100(2), 2019.
\newblock Cited by: 54; All Open Access, Bronze Open Access.

\bibitem{weak-sindy}
Daniel~A. Messenger and David~M. Bortz.
\newblock Weak sindy for partial differential equations.
\newblock {\em Journal of Computational Physics}, 443:110525, 2021.

\bibitem{reactive-sindy}
Moritz Hoffmann, Christoph Fröhner, and Frank Noé.
\newblock {Reactive SINDy: Discovering governing reactions from concentration data}.
\newblock {\em The Journal of Chemical Physics}, 150(2):025101, 01 2019.

\bibitem{time_delay_discrete_discovery}
Alessandro Pecile, Nicola Demo, Marco Tezzele, Gianluigi Rozza, and Dimitri Breda.
\newblock Data-driven discovery of delay differential equations with discrete delays.
\newblock page null, 07 2024.

\bibitem{mho-discovery}
Fernando Lejarza and Michael Baldea.
\newblock Data-driven discovery of the governing equations of dynamical systems via moving horizon optimization.
\newblock {\em Scientific Reports}, 12(1):11836, Jul 2022.

\bibitem{FUENTES2021107528}
R.~Fuentes, R.~Nayek, P.~Gardner, N.~Dervilis, T.~Rogers, K.~Worden, and E.J. Cross.
\newblock Equation discovery for nonlinear dynamical systems: A bayesian viewpoint.
\newblock {\em Mechanical Systems and Signal Processing}, 154:107528, 2021.

\bibitem{nayek2021spike}
R.~Nayek, R.~Fuentes, K.~Worden, and E.J. Cross.
\newblock On spike-and-slab priors for bayesian equation discovery of nonlinear dynamical systems via sparse linear regression.
\newblock {\em Mechanical Systems and Signal Processing}, 161:107986, 2021.

\bibitem{gupta2021bayesian}
Kushagra Gupta, Dootika Vats, and Snigdhansu Chatterjee.
\newblock Bayesian equation selection on sparse data for discovery of stochastic dynamical systems.
\newblock {\em arXiv preprint arXiv:2101.04437}, 2021.

\bibitem{gibbs-stochastic}
Tapas Tripura and Souvik Chakraborty.
\newblock A sparse bayesian framework for discovering interpretable nonlinear stochastic dynamical systems with gaussian white noise.
\newblock {\em Mechanical Systems and Signal Processing}, 187:109939, 2023.

\bibitem{nayek2022equation}
Rajdip Nayek, Keith Worden, and Elizabeth~J Cross.
\newblock Equation discovery using an efficient variational bayesian approach with spike-and-slab priors.
\newblock In {\em Model Validation and Uncertainty Quantification, Volume 3}, pages 149--161. Springer, 2022.

\bibitem{bayesian_variable_selection}
T.~J. Mitchell and J.~J. Beauchamp.
\newblock Bayesian variable selection in linear regression.
\newblock {\em Journal of the American Statistical Association}, 83(404):1023--1032, 1988.

\bibitem{gibbs-variable-selection}
Edward~I. George and Robert~E. McCulloch.
\newblock Variable selection via gibbs sampling.
\newblock {\em Journal of the American Statistical Association}, 88(423):881--889, 1993.

\bibitem{ss-prior-frequentist}
Hemant Ishwaran and J.~Sunil Rao.
\newblock {Spike and slab variable selection: Frequentist and Bayesian strategies}.
\newblock {\em The Annals of Statistics}, 33(2):730 -- 773, 2005.

\bibitem{spike-slab-variable-selection}
Naveen~Naidu Narisetty and Xuming He.
\newblock Bayesian variable selection with shrinking and diffusing priors.
\newblock {\em The Annals of Statistics}, 42(2):789--817, 2014.

\bibitem{hirsh2021sparsifying}
Seth~M. Hirsh, David~A. Barajas-Solano, and J.~Nathan Kutz.
\newblock Sparsifying priors for bayesian uncertainty quantification in model discovery.
\newblock 2021.

\bibitem{mangan2017model}
Niall~M Mangan, J~Nathan Kutz, Steven~L Brunton, and Joshua~L Proctor.
\newblock Model selection for dynamical systems via sparse regression and information criteria.
\newblock {\em Proceedings of the Royal Society A: Mathematical, Physical and Engineering Sciences}, 473(2204):20170009, 2017.

\bibitem{ROSAFALCO}
Luca Rosafalco, Paolo Conti, Andrea Manzoni, Stefano Mariani, and Attilio Frangi.
\newblock Ekf–sindy: Empowering the extended kalman filter with sparse identification of nonlinear dynamics.
\newblock {\em Computer Methods in Applied Mechanics and Engineering}, 431:117264, 2024.

\bibitem{accel_gaussian_process}
Ben Calderhead, Mark Girolami, and Neil Lawrence.
\newblock Accelerating bayesian inference over nonlinear differential equations with gaussian processes.
\newblock In D.~Koller, D.~Schuurmans, Y.~Bengio, and L.~Bottou, editors, {\em Advances in Neural Information Processing Systems}, volume~21. Curran Associates, Inc., 2008.

\bibitem{bio-distributed-delays}
Boseung Choi, Yu-Yu Cheng, Selahattin Cinar, William Ott, Matthew~R Bennett, Krešimir Josić, and Jae~Kyoung Kim.
\newblock {Bayesian inference of distributed time delay in transcriptional and translational regulation}.
\newblock {\em Bioinformatics}, 36(2):586--593, 07 2019.

\bibitem{WU2023133647}
Yuqiang Wu.
\newblock Reconstruction of delay differential equations via learning parameterized dictionary.
\newblock {\em Physica D: Nonlinear Phenomena}, 446:133647, 2023.

\bibitem{time_delay_gap_effects_sindy}
Jiamin Xu, Nazli Demirer, Vy~Pho, Kaixiao Tian, He~Zhang, Ketan Bhaidasna, Robert Darbe, and Dongmei Chen.
\newblock Data-driven modeling of nonlinear delay differential equations with gap effects using sindy.
\newblock In {\em 2024 IEEE International Conference on Advanced Intelligent Mechatronics (AIM)}, pages 198--203, 2024.

\bibitem{Sandoz_2023}
Antoine Sandoz, Verena Ducret, Georg~A. Gottwald, Gilles Vilmart, and Karl Perron.
\newblock Sindy for delay-differential equations: application to model bacterial zinc response.
\newblock {\em Proceedings of the Royal Society A: Mathematical, Physical and Engineering Sciences}, 479(2269), January 2023.

\bibitem{Leylaz_2021}
Ghazaale Leylaz, Ghazaale Leylaz, Shuo Wang, Shuo Wang, Jian‐Qiao Sun, and Jian-Qiao Sun.
\newblock Identification of nonlinear dynamical systems with time delay.
\newblock {\em International Journal of Dynamics and Control}, 2021.

\bibitem{data-driven-augment}
{\'A}kos Tam{\'a}s K{\"o}peczi-B{\'o}cz, Henrik Sykora, and D{\'e}nes Tak{\'a}cs.
\newblock Data-driven delay identification with sindy.
\newblock In Walter Lacarbonara, editor, {\em Advances in Nonlinear Dynamics, Volume III}, pages 481--491, Cham, 2024. Springer Nature Switzerland.

\bibitem{JCSPROTT}
J.C. Sprott.
\newblock A simple chaotic delay differential equation.
\newblock {\em Physics Letters A}, 366(4):397--402, 2007.

\bibitem{mackey-glass}
Michael~C. Mackey and Leon Glass.
\newblock Oscillation and chaos in physiological control systems.
\newblock {\em Science}, 197(4300):287--289, 1977.

\bibitem{brunton2016discovering}
Steven~L Brunton, Joshua~L Proctor, and J~Nathan Kutz.
\newblock Discovering governing equations from data by sparse identification of nonlinear dynamical systems.
\newblock {\em Proceedings of the national academy of sciences}, 113(15):3932--3937, 2016.

\end{thebibliography}

\end{document}